\documentclass[10pt,twocolumn,letterpaper]{article}

\usepackage[pagenumbers]{cvpr} % To force page numbers, e.g. for an arXiv version
\usepackage[acronym,toc]{glossaries}
\glsdisablehyper

\newacronym{vla}{VLA}{Vision-Language-Action model}
\newacronym{vlm}{VLM}{Vision-Language Model}
\newacronym{wam}{WAM}{World-Action Model}
\newacronym{ood}{OOD}{out-of-distribution}
\newacronym{paiav}{PAI-AV}{PhysicalAI-Autonomous-Vehicles}
\newacronym{reactnav}{ReactNAV}{NAVSIM}
\newacronym{reactpai}{ReactPAI}{PAI-AV}
\newacronym{vqa}{VQA}{Visual Question Answering}
\newacronym{sota}{SOTA}{state-of-the-art}
\newacronym{llm}{LLM}{Large Language Model}
\newacronym{moe}{MoE}{Mixture of Experts}
\newacronym{ade}{ADE}{Average Displacement Error}
\newacronym{cot}{CoT}{Chain-of-Thought}
\newacronym{pdms}{PDMS}{Predictive Driver Model Score}
\newacronym{sft}{SFT}{supervised fine-tuning}

\glsaddall
\usepackage{algorithm}
\usepackage{algpseudocode}
\usepackage{mathtools}
\usepackage{booktabs}
\usepackage{multirow}
\usepackage{array,tabularx,makecell}

\usepackage{colortbl}

\usepackage{microtype}

\newcommand{\benchname}{\mbox{ReactBench-AD}\xspace}

\newcommand{\dP}{\ensuremath{\Delta P}}
\newcommand{\NC}{\ensuremath{\mathrm{NC}}}
\newcommand{\NCins}{\ensuremath{\mathrm{NC}_{\mathrm{ins}}}}
\newcommand{\dNCins}{\ensuremath{\Delta\mathrm{NC}_{\mathrm{ins}}}}
\newcommand{\DAC}{\ensuremath{\mathrm{DAC}}}
\newcommand{\CSC}{\ensuremath{\mathrm{\Delta CS}}}
\newcommand{\CS}{\ensuremath{\mathrm{CS}}}
\newcommand{\TIR}{\ensuremath{\mathrm{TIR}_{\epsilon}}}

\definecolor{cvprblue}{rgb}{0.21,0.49,0.74}
\usepackage[pagebackref,breaklinks,colorlinks,allcolors=cvprblue]{hyperref}

\title{\benchname: Do Driving VLAs and World-Action Models React to Imminent Hazards in Their Path?}

\author{Nikos Theodoridis$^{1}$\thanks{Corresponding author: \texttt{theodoridis.nikolaos@ul.ie}} \quad Reenu Mohandas$^{1}$ \quad Sushil Sharma$^{1}$ \quad Ciar\'{a}n Hogan$^{2}$\\
Ganesh Sistu$^{1,2}$ \quad Anthony Scanlan$^{1}$ \quad Tim Brophy$^{1}$ \quad Ciar\'{a}n Eising$^{1}$\\[0.3em]
$^{1}$University of Limerick \qquad $^{2}$Valeo Vision Systems
}

\begin{document}
\maketitle
\begin{abstract}
    Vision-Language-Action models (VLAs) and World-Action Models (WAMs) have emerged in autonomous driving with the promise of strong generalization to long-tail scenarios. Although these models achieve state-of-the-art results on autonomous driving benchmarks, most existing benchmarks are dominated by nominal scenarios and therefore cannot test this promise. To test one such long-tail capability that is critical for safety, we introduce \textbf{\benchname}, a benchmark that measures how driving models react to objects that appear in the ego vehicle's path. Built on real NAVSIM and PhysicalAI-Autonomous-Vehicles scenes, \benchname composites a crossing pedestrian, a standing pedestrian, or a stationary car consistently across all camera views and history frames. Each edited scene thus forms an exact, pixel-level counterfactual pair with its original, which allows us to isolate the causal effect of the inserted object on the predicted trajectory. Evaluating four VLAs and two WAMs, we find that they react weakly, often barely at all. Finally, we test how easily one of the evaluated VLAs, ReCogDrive, can be taught to brake hard for imminent hazards: a short fine-tuning stage on edited scenes substantially improves its reactions, with the learned behavior generalizing to unseen objects and fully real scenes, at minimal cost to nominal driving. These results call into question the out-of-the-box long-tail capabilities of these models, but suggest that they can be a strong foundation for acquiring them with little additional training, highlighting the value of real or synthetic data that demonstrates the long-tail behaviors nominal datasets rarely contain. Code is available at \url{https://github.com/niktheod/reactbench-ad}.
\end{abstract}    
\begin{figure}[t]
    \centering
    \includegraphics[width=\linewidth]{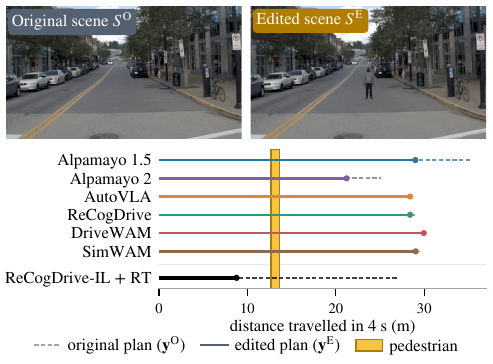}

    \caption{We insert an object, here a standing pedestrian, into the ego vehicle's path in a real driving scene, so that the edited scene $S^{\mathrm{E}}$ and the original scene $S^{\mathrm{O}}$ differ only in the pixels of the inserted object. Bottom: distance travelled by each model's front bumper over its 4\,s plan in the original ($\mathbf{y}^{\mathrm{O}}$) and edited ($\mathbf{y}^{\mathrm{E}}$) scene, with the pedestrian's extent shown in yellow. All six evaluated \acrshortpl{vla} and \acrshortpl{wam} shorten their plans only slightly, if at all, and drive through the pedestrian, whereas the ReCogDrive IL checkpoint stops before it after a short fine-tuning on edited scenes (IL + RT, \cref{sec:teaching_to_react}).}
    \label{fig:main}
\end{figure}

\section{Introduction}
\glspl{vla} build on top of the capabilities of \glspl{vlm}, steering them from general purpose chat models to policy networks that can output low-level actions for the manipulation of robots or vehicles~\cite{pmlr-v229-zitkovich23a, xu2024drivegpt4, hwang2025emma, li2026recogdrive, wang2025alpamayo}. The main motivation for using them in autonomous driving is that they can inherit the reasoning and generalization capabilities of \glspl{vlm} to better handle long-tail and \gls{ood} scenarios, which modular pipelines struggle with~\cite{hwang2025emma}. Indeed, these models show promising results, with many of them~\cite{li2026recogdrive, zhou2025autovla, Renz_2025_CVPR, Fu_2025_ICCV, tian2024drivevlm, wang2025omnidrive, hwang2025emma} topping open- and closed-loop benchmarks such as NAVSIM~\cite{NEURIPS2024_32768f7f}, NAVSIMv2~\cite{cao2025pseudosimulation}, Bench2Drive~\cite{NEURIPS2024_017761f9}, and nuScenes~\cite{9156412}.

A newer line of work uses a different type of foundation model to build on top of, namely, video generation models. By fine-tuning these models to denoise action tokens along with video tokens they are turned into what has recently been termed \glspl{wam}~\cite{ye2026world}. The main argument as to why \glspl{wam} might be more adequate for embodied AI applications is that they can take advantage of the dynamic representations of how the world evolves that they inherit from their backbone video generation model. Likewise, some of these models~\cite{shi2026drivewam, Xia_2026_CVPR, fu20264d, liu2026universe, zhao2026simwam} have demonstrated \gls{sota} performance on multiple autonomous driving benchmarks.

However, the vast majority of the samples that compose these benchmarks are nominal driving scenarios where unexpected events don't usually occur, so how these models react to safety-critical scenarios, and in particular to imminent hazards in their path, remains understudied. Evaluating this reliably places requirements both on the inputs the models see and on how their reactions are measured. On the input side, the scenes should ideally be real rather than simulated, and they should match the input configuration each model was trained on. On the measurement side, each safety-critical scene should be paired with an otherwise identical counterpart that lacks the hazardous object, so that the object's causal effect on the predicted trajectory can be isolated. For such a pair to be informative, the object must be placed where a reaction is actually required, i.e., on the ego vehicle's path, and it must appear consistently in every camera view and history frame whose field of view covers its location. In addition, the metrics should describe how a model reacts, not only whether it collides: aggregate collision rates cannot distinguish a model that ignores a hazard from one that reacts, but too weakly or too late. To our knowledge no benchmark meets all of these requirements at once.

To address this gap, we introduce \textbf{\benchname}, a benchmark for evaluating how \glspl{vla} and \glspl{wam} react to objects inserted in their path (\cref{fig:main}). Since many \glspl{vla} and \glspl{wam} are trained on NAVSIM~\cite{NEURIPS2024_32768f7f} or \gls{paiav}~\cite{nvidia2025paiav}, we build \benchname on these two datasets, so that these models are evaluated on inputs matching their training configuration, with the same cameras, calibrations, history length, and frame rate. Into each sample, we separately insert a crossing pedestrian, a standing pedestrian, and a stationary car at a controlled position on the ego vehicle's path, rendering the object consistently across all camera views and history frames and modifying only the pixels of the object and its shadow. Each edited sample thus forms an exact counterfactual pair with its original, so any change in the predicted trajectory is caused by the inserted object.

We evaluate four \glspl{vla}, namely Alpamayo~1.5~\cite{nvidia2026alpamayo15}, Alpamayo~2~\cite{nvidia2026alpamayo2}, AutoVLA~\cite{zhou2025autovla}, and ReCogDrive~\cite{li2026recogdrive}, and two \glspl{wam}, namely DriveWAM~\cite{shi2026drivewam} and SimWAM~\cite{zhao2026simwam}, and find that they react weakly to inserted objects and fail to avoid the collision in most cases. To understand why, we examine two potential factors: \gls{cot} reasoning and the individual stages of the multi-stage training recipes that some of these models follow. For inserted pedestrians, the Alpamayo models recognize the hazard in their CoT but mostly fail to act on it. Finally, a short fine-tuning stage on edited scenes teaches ReCogDrive to brake hard, generalizing to unseen objects and fully real scenes. Together, these findings suggest that the models often recognize hazards but lack the behaviors to act on them.

Our contributions can be summarized as follows.

\begin{itemize}
    \item We introduce \benchname, a benchmark of exact, pixel-level counterfactual pairs built on real NAVSIM and \gls{paiav} scenes, with metrics that capture whether and how models react to objects in their path.
    \item We evaluate four \glspl{vla} and two \glspl{wam} and show that they largely fail to handle such imminent hazards.
    \item We investigate how CoT reasoning and the individual training stages affect their reactions.
    \item We show that fine-tuning only ReCogDrive's action head on edited scenes teaches it to brake for hazards, including unseen objects and fully real scenes, at almost no cost to nominal driving.
\end{itemize}

\begin{figure*}[t]
    \centering
    \includegraphics[width=\linewidth]{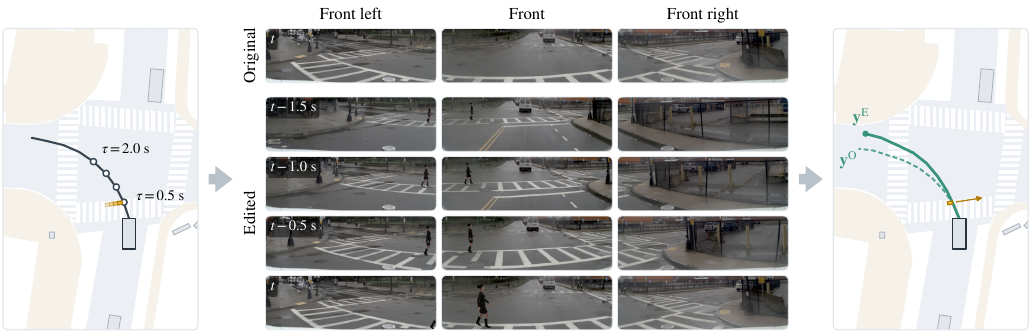}

    \caption{Overview of \benchname. Left: we pick a target on the ego vehicle's ground-truth trajectory (dark line), namely the ego's front-bumper position at a look-ahead time $\tau \in \{0.5, 1.0, 1.5, 2.0\}\,\mathrm{s}$ after the current frame $t$ (circles). At each history timestep, a standing object is placed at the target, whereas a moving one, such as the crossing pedestrian shown here, is placed along a path that reaches the target at $t + \tau$ (here $\tau = 0.5\,\mathrm{s}$). Middle: we composite the object at these positions into every camera view and history frame of the model's input (top: the original scene at the current frame $t$). Right: in the edited scene, we score the model's plans for the original ($\mathbf{y}^{\mathrm{O}}$, dashed) and edited ($\mathbf{y}^{\mathrm{E}}$, solid) input (\cref{eq:edit_effect}), here for ReCogDrive.}
    \label{fig:overview}
\end{figure*}

\section{Related Work}

\paragraph{VLAs and WAMs for Autonomous Driving.} 
\glspl{vla} fine-tune pretrained \glspl{vlm} to predict robot actions in the language space~\cite{pmlr-v229-zitkovich23a}, an idea quickly adopted for predicting driving trajectories or low-level actions for autonomous driving~\cite{xu2024drivegpt4, tian2024drivevlm, wang2025omnidrive, hwang2025emma}. Later \glspl{vla} moved action prediction out of the language space, using special action tokens~\cite{zhou2025autovla} or a dedicated action head, ranging from MLPs~\cite{Shao_2024_CVPR, Renz_2025_CVPR} to generative models~\cite{Fu_2025_ICCV, li2026recogdrive, wang2025alpamayo}. \glspl{wam} have recently emerged as an alternative to \glspl{vla}, replacing the \gls{vlm} backbone with a pretrained video generation model~\cite{ye2026world, PaiJ-RSS-26}. Driving \glspl{wam} either denoise action tokens conditioned on (partially) denoised video tokens~\cite{shi2026drivewam, Xia_2026_CVPR}, denoise both jointly~\cite{fu20264d}, or use future video prediction only as an auxiliary training objective, avoiding frame generation at inference~\cite{liu2026universe, zhao2026simwam}. Although many of these models achieve \gls{sota} results on open- and closed-loop benchmarks, their ability to handle safety-critical scenarios remains understudied.

\paragraph{Safety Evaluation of End-to-End Planners.}
Several benchmarks evaluate driving models in safety-critical scenarios. SafeBench~\cite{xu2022safebench} and Bench2Drive~\cite{NEURIPS2024_017761f9} provide such scenarios in the CARLA simulator~\cite{pmlr-v78-dosovitskiy17a}, the latter as part of a broader multi-ability evaluation, while Fail2Drive~\cite{gerstenecker2026fail2drive} pairs CARLA routes containing distribution shifts, such as unseen obstacles, with in-distribution counterparts and reports consistent performance drops. However, since CARLA renders synthetic images, planners trained on real-world data, such as the \glspl{vla} and \glspl{wam} we study, cannot be evaluated there without a large sim-to-real gap. WOD-E2E~\cite{Xu_2026_CVPR} curates real long-tail driving scenes for open-loop evaluation, selected for rarity rather than for requiring an immediate reaction, and interPlan~\cite{10803052} augments nuPlan scenarios with long-tail events such as jaywalkers and accident sites, but only at the object level, so it cannot evaluate sensor-based planners.

To keep sensor inputs realistic, another line of work builds safety-critical scenarios on top of real driving scenes using neural reconstruction or generative models. NeuroNCAP~\cite{10.1007/978-3-031-73404-5_10} builds a closed-loop safety testing framework on NeRF~\cite{10.1145/3503250, 10657387} reconstructions of nuScenes scenes, and HUGSIM~\cite{11314796} builds a closed-loop simulator on 3D Gaussian Splatting~\cite{kerbl:hal-04088161} reconstructions of nuScenes and other datasets, whose harder scenarios include aggressive vehicles. Challenger~\cite{xu2025challenger} and SMD~\cite{pmlr-v306-zhou26aj} instead generate multi-view nuScenes videos of safety-critical scenarios, while Geng et al.~\cite{geng2025driving} extend generative rendering to closed-loop evaluation with adversarial vehicles. Since many open-weight \glspl{vla} and \glspl{wam}, including all models we evaluate, are trained on NAVSIM or \gls{paiav}, evaluating these models on such benchmarks would confound their failures to react with sensor and domain shift. RealEngine~\cite{jiang2025realengine} and NavSafe~\cite{bao2026navsafe} reconstruct nuPlan scenes, matching the training domain of NAVSIM-based models, but imminent hazards requiring a prompt reaction are only a subset of their scenarios. Since all of these methods re-render or synthesize the entire camera image, none of them yields exact counterfactual pairs with the original sensor data, whereas \benchname modifies only the inserted object's pixels, isolating its causal effect on the predicted trajectory.

\paragraph{Counterfactual Safety Evaluation.}
Closest to our work are two contemporaneous studies that edit real scenes to measure the causal effect of objects on predicted trajectories. Yang et al.~\cite{yang2026beyond} inpaint away pedestrians near the ego corridor in nuScenes and NAVSIM frames and find that end-to-end and \gls{vla} planners rarely exhibit genuine avoidance. Luo et al.~\cite{luo2026exceptiondrive} insert hazards into nuScenes views via instruction-based image editing and show that planners frequently drive into them. Our work differs from these studies in four ways. First, \benchname places the object exactly on the ego vehicle's ground-truth future trajectory, so it always obstructs the path the vehicle actually drove. Second, it uses the camera calibration to render the object into every camera view and history frame, leaving all other pixels unchanged. Third, we evaluate every model on scenes that match its training distribution, so that performance drops caused by domain shift are not mistaken for failures in safety-critical scenes. Fourth, we also evaluate \glspl{wam}.

\section{\benchname}
\label{sec:reactbench}

\benchname pairs real driving scenes with edited counterparts in which an object blocks the ego vehicle's path at close range, so that a prompt reaction, either braking or a safe nudge around the object, is needed to avoid a collision.

\subsection{Object Compositing}
\label{subsec:object_compositing}
To keep the inserted object consistent across camera views and timesteps, control its 3D position precisely, and modify only the pixels of the object and its shadow, we render 3D assets with Blender~\cite{blender} and composite them into the original scenes. For each scene, we generate three counterfactual variants: a crossing pedestrian, a standing pedestrian, and a stationary car, drawn from 42 human and seven vehicle assets. Each object is placed at the ego vehicle's front-bumper position on its ground-truth trajectory after a look-ahead time $\tau \in \{0.5, 1.0, 1.5, 2.0\}\,\mathrm{s}$, so that, had the ego vehicle kept driving as logged, it would collide with the object within $\tau$. The crossing pedestrian instead walks toward the ego vehicle's path at 1.4\,m/s and reaches this point after $\tau$ (\cref{fig:overview}). We render the object, with a contact shadow, into the four history timesteps and up to four camera views and hide the parts occluded by existing scene geometry. The inserted objects are clearly recognizable: an off-the-shelf COCO detector (D-FINE-X~\cite{ICLR2025_6cf58a87}) finds them in 99.3\% of the edited front-camera images (Appendix~\ref{subsec:supp_detection_check}); further details and visualizations are provided in Appendix~\ref{subsec:supp_object_compositing}.

\subsection{Data and Evaluation Protocol}
\label{subsec:benchmark_details}

\paragraph{Sample Selection.}
\benchname is constructed from samples drawn from \gls{reactnav} and \gls{reactpai}. For \gls{reactnav}, we filter navtest, the test split of NAVSIM, to exclude samples in which the ego vehicle is nearly stationary (below 5\,km/h at the current frame $t$), as these require no reaction. We then keep only samples in which each of the three objects---a crossing pedestrian, a standing pedestrian, and a stationary car---can be placed at all four history timesteps up to and including $t$. This filtering yields 3,228 navtest samples, which form \gls{reactnav}. For \gls{reactpai}, we randomly select and download 5,500 clips from \gls{paiav} and apply the same filters, retaining 1,045 samples. In both subsets, the ego-vehicle speed at the current frame lies between 5 and 40\,km/h (Appendix~\ref{subsec:supp_data_stats}).

\paragraph{Metrics.}
Let $S^{\mathrm{O}}$ denote the original scene and $S^{\mathrm{E}}$ its edited, asset-augmented counterpart, and let $\pi_\theta$ be the evaluated model. We obtain two trajectories, $\mathbf{y}^{\mathrm{O}} = \pi_\theta(S^{\mathrm{O}})$ and $\mathbf{y}^{\mathrm{E}} = \pi_\theta(S^{\mathrm{E}})$, predicted from the original and edited scene, respectively. For a metric $\mathcal{M}$, we write $\mathcal{M}(\mathbf{y}; S)$ for its value when trajectory $\mathbf{y}$ is evaluated in scene $S$, and define the edit effect as
\begin{equation}
    \Delta \mathcal{M} = \mathcal{M}^{\mathrm{E}} - \mathcal{M}^{\mathrm{O}} = \mathcal{M}(\mathbf{y}^{\mathrm{E}}; S^{\mathrm{E}}) - \mathcal{M}(\mathbf{y}^{\mathrm{O}}; S^{\mathrm{E}}).
    \label{eq:edit_effect}
\end{equation}
Note that both trajectories are evaluated in the edited scene $S^{\mathrm{E}}$: $\mathbf{y}^{\mathrm{O}}$ is scored against the inserted object even though the model never observed it. Hence, $\mathcal{M}^{\mathrm{O}}$ reflects the outcome had the model ignored the hazard, and $\Delta \mathcal{M}$ isolates the effect of the model's reaction to it.

For \gls{reactnav}, our primary metric is ${P \coloneqq \NC \times \DAC}$, which combines the no-collision (at-fault) rate, $\NC$, with drivable area compliance, $\DAC$; for \gls{reactpai}, since \gls{paiav} lacks the map data needed for $\DAC$, we use ${P \coloneqq \NC}$. To capture reactions that improve safety without avoiding a collision, we also use the mean contact speed, $\CS$, over the samples for which both $\mathbf{y}^{\mathrm{O}}$ and $\mathbf{y}^{\mathrm{E}}$ result in an at-fault collision. For both metrics, we report the edit effect (\cref{eq:edit_effect}), $\dP$ and $\CSC$. Finally, to assess whether the models respond to the inserted actors at all, successfully or not, we report the \emph{trajectory invariance rate}, $\TIR$, defined as the percentage of samples for which the \gls{ade} between $\mathbf{y}^{\mathrm{O}}$ and $\mathbf{y}^{\mathrm{E}}$ is below a threshold $\epsilon$.

We follow the at-fault criteria of NAVSIM with three small modifications that make them stricter, e.g., a lateral contact with a pedestrian or cyclist counts as at fault. Additionally, since driving models typically predict waypoints at a low temporal resolution (e.g., 2\,Hz), we interpolate each predicted trajectory piecewise-linearly and check collisions at 100\,Hz, so that contacts between waypoints are neither missed nor misclassified. Details on the at-fault criteria and the interpolation are provided in Appendix~\ref{subsec:supp_metrics_trajectory}.

\section{Experiments and Results}
\label{sec:experiments}

\subsection{Experimental Setup}
\label{subsec:models}
We evaluate four \glspl{vla}, Alpamayo~1.5~\cite{nvidia2026alpamayo15}, Alpamayo~2~\cite{nvidia2026alpamayo2}, AutoVLA~\cite{zhou2025autovla}, and ReCogDrive~\cite{li2026recogdrive} (2B variant), and two \glspl{wam}, DriveWAM~\cite{shi2026drivewam} and SimWAM~\cite{zhao2026simwam}, whose architectures and input configurations are detailed in Appendix~\ref{subsec:supp_models}. ReCogDrive, DriveWAM, and SimWAM receive only the current front-camera frame, whereas AutoVLA and the Alpamayo models receive several cameras and history frames. AutoVLA, ReCogDrive, DriveWAM, and SimWAM were trained on NAVSIM and are evaluated on \gls{reactnav} only, whereas the Alpamayo models were trained primarily on NVIDIA's multi-camera driving data, including \gls{paiav}, and are evaluated on both subsets. On \gls{reactnav}, the Alpamayo models receive only the three front-facing cameras, with frames 0.5\,s instead of 0.1\,s apart, and hence operate under a strong input distribution shift, which should be kept in mind when interpreting their results. Finally, since the models' prediction horizons range from 4.0 to 6.4\,s, we compute all metrics on only the first 4\,s of each predicted trajectory to compare all models over the same horizon.

\subsection{Results}
\label{subsec:eval_results}
 \Cref{tab:main_results} summarizes the main results, with more detailed results along with trajectory visualizations provided in Appendix~\ref{subsec:supp_results}. In most cases, the models fail to meaningfully improve the safety score $P$ when the inserted object is visible compared to when it is not ($\mathbf{y}^{\mathrm{E}}$ vs.\ $\mathbf{y}^{\mathrm{O}}$). The main exceptions are the two Alpamayo models, which achieve substantial improvements at longer look-ahead times on both \gls{reactnav} and \gls{reactpai}. Nevertheless, their performance remains far from ideal: even in the best case, an inserted standing pedestrian at $\tau = 2.0\,\mathrm{s}$ on \gls{reactpai} for Alpamayo~1.5, unsafe trajectories remain the most likely outcome, accounting for 62.01\% of cases. Excluding the samples in which even braking at $8\,\mathrm{m/s^2}$~\cite{greibe2007braking,8813885} cannot prevent a collision barely changes $\dP$ (Appendix~\ref{subsec:supp_unavoidable}).

\begin{table*}[t]
    \centering
    \caption{Results on ReactNAV and ReactPAI by model, inserted object, and look-ahead time $\tau$ (s). $\dP$ is in percentage points (higher is better); $\CSC$ is in km/h (lower is better). Gray: the paired 95\% bootstrap CI includes zero.}
    \label{tab:main_results}
    \footnotesize
    \setlength{\tabcolsep}{4.1pt}
    \begin{tabular}{ll*{4}{r}@{\hspace{16pt}}*{4}{r}@{\hspace{16pt}}*{4}{r}}
        \toprule
        & & \multicolumn{4}{c}{Crossing pedestrian} & \multicolumn{4}{c}{Standing pedestrian} & \multicolumn{4}{c}{Car} \\
        \cmidrule(lr){3-6} \cmidrule(lr){7-10} \cmidrule(lr){11-14}
        Model & Metric & $\tau{=}0.5$ & 1.0 & 1.5 & 2.0 & 0.5 & 1.0 & 1.5 & 2.0 & 0.5 & 1.0 & 1.5 & 2.0 \\
        \midrule
        \multicolumn{14}{l}{\textit{ReactNAV}} \\
        \midrule
        \rowcolor{gray!15} \cellcolor{white}Alpamayo~1.5 & $\boldsymbol{\dP}$ & $+0.25$ & $+3.10$ & $+7.16$ & $+8.71$ & \textcolor{gray}{$+0.09$} & $+3.66$ & $+11.93$ & $+17.63$ & $+0.12$ & $+3.84$ & $+15.92$ & $+28.81$ \\
         & $\CSC$ & $-1.51$ & $-2.73$ & $-2.68$ & $-2.29$ & $-1.63$ & $-4.52$ & $-5.44$ & $-4.83$ & $-1.23$ & $-4.29$ & $-7.05$ & $-7.25$ \\
        \addlinespace[2pt]
        \rowcolor{gray!15} \cellcolor{white}Alpamayo~2 & $\boldsymbol{\dP}$ & \textcolor{gray}{$+0.06$} & $+2.11$ & $+6.75$ & $+8.36$ & \textcolor{gray}{$+0.03$} & $+3.25$ & $+9.42$ & $+12.33$ & $+0.12$ & $+5.36$ & $+19.39$ & $+22.43$ \\
         & $\CSC$ & $-1.11$ & $-1.77$ & $-1.42$ & $-0.73$ & $-1.49$ & $-3.10$ & $-3.18$ & $-2.50$ & $-1.19$ & $-2.96$ & $-4.17$ & $-4.22$ \\
        \addlinespace[2pt]
        \rowcolor{gray!15} \cellcolor{white}AutoVLA & $\boldsymbol{\dP}$ & \textcolor{gray}{$+0.00$} & \textcolor{gray}{$-0.03$} & $+0.53$ & $+0.90$ & \textcolor{gray}{$+0.00$} & $+0.16$ & $+0.34$ & $+0.96$ & \textcolor{gray}{$+0.03$} & $+0.87$ & $+2.14$ & $+3.26$ \\
        & $\CSC$ & $+0.03$ & \textcolor{gray}{$-0.02$} & \textcolor{gray}{$-0.02$} & \textcolor{gray}{$-0.03$} & \textcolor{gray}{$-0.01$} & $-0.08$ & $-0.16$ & $-0.16$ & $-0.28$ & $-0.59$ & $-0.70$ & $-0.67$ \\
        \addlinespace[2pt]
        \rowcolor{gray!15} \cellcolor{white}ReCogDrive & $\boldsymbol{\dP}$ & \textcolor{gray}{$+0.00$} & \textcolor{gray}{$-0.03$} & $+0.22$ & $+0.40$ & \textcolor{gray}{$+0.00$} & \textcolor{gray}{$+0.00$} & $+0.12$ & $+0.40$ & \textcolor{gray}{$+0.00$} & \textcolor{gray}{$+0.00$} & $+0.28$ & $+1.89$ \\
         & $\CSC$ & $+0.31$ & $-0.45$ & $-0.57$ & $-0.45$ & $+0.51$ & $-0.26$ & $-1.06$ & $-1.20$ & $+1.42$ & $-0.07$ & $-0.96$ & $-1.08$ \\
        \addlinespace[2pt]
        \rowcolor{gray!15} \cellcolor{white}DriveWAM & $\boldsymbol{\dP}$ & \textcolor{gray}{$+0.00$} & \textcolor{gray}{$+0.06$} & $+0.12$ & \textcolor{gray}{$+0.09$} & \textcolor{gray}{$+0.00$} & $+0.12$ & $+0.56$ & $+2.08$ & \textcolor{gray}{$+0.00$} & $+1.77$ & $+3.62$ & $+5.61$ \\
         & $\CSC$ & $-0.08$ & $-0.14$ & $-0.17$ & $-0.14$ & $-0.15$ & $-0.43$ & $-0.72$ & $-0.88$ & $-0.60$ & $-0.88$ & $-0.91$ & $-0.76$ \\
        \addlinespace[2pt]
        \rowcolor{gray!15} \cellcolor{white}SimWAM & $\boldsymbol{\dP}$ & \textcolor{gray}{$+0.00$} & \textcolor{gray}{$+0.06$} & $+0.12$ & $+0.22$ & \textcolor{gray}{$+0.00$} & \textcolor{gray}{$+0.00$} & $+0.15$ & $+0.71$ & \textcolor{gray}{$+0.00$} & \textcolor{gray}{$+0.06$} & $+0.99$ & $+2.26$ \\
         & $\CSC$ & $-0.11$ & $-0.14$ & $-0.82$ & $-0.32$ & $-0.07$ & $-0.21$ & $-1.64$ & $-1.26$ & $+0.45$ & $-0.79$ & $-1.79$ & $-0.79$ \\
        \midrule
        \multicolumn{14}{l}{\textit{ReactPAI}} \\
        \midrule
        \rowcolor{gray!15} \cellcolor{white}Alpamayo~1.5 & $\boldsymbol{\dP}$ & \textcolor{gray}{$+0.10$} & $+5.55$ & $+25.17$ & $+35.12$ & \textcolor{gray}{$+0.19$} & $+5.07$ & $+21.63$ & $+35.69$ & $+0.67$ & $+12.92$ & $+20.86$ & $+25.65$ \\
         & $\CSC$ & $-3.64$ & $-7.96$ & $-7.52$ & $-5.75$ & $-4.13$ & $-9.14$ & $-10.25$ & $-8.79$ & $-4.43$ & $-9.08$ & $-9.73$ & $-8.52$ \\
        \addlinespace[2pt]
        \rowcolor{gray!15} \cellcolor{white}Alpamayo~2 & $\boldsymbol{\dP}$ & \textcolor{gray}{$+0.00$} & $+3.44$ & $+22.68$ & $+34.07$ & \textcolor{gray}{$+0.00$} & $+0.57$ & $+8.61$ & $+15.89$ & \textcolor{gray}{$+0.10$} & $+5.55$ & $+16.65$ & $+25.17$ \\
         & $\CSC$ & $-1.80$ & $-6.01$ & $-6.69$ & $-5.42$ & $-2.45$ & $-5.85$ & $-6.31$ & $-5.93$ & $-3.07$ & $-8.16$ & $-8.49$ & $-6.95$ \\
        \bottomrule
    \end{tabular}
\end{table*}

On \gls{reactnav}, $\dP$ is smallest for crossing pedestrians, larger for standing pedestrians, and largest for stationary cars. This ordering holds for almost all evaluated models, although the magnitudes differ considerably. For AutoVLA, ReCogDrive, DriveWAM, and SimWAM, improvements exceeding 1 percentage point occur only for stationary cars, with the sole exception of DriveWAM, which also improves slightly for the standing pedestrian at $\tau = 2.0\,\mathrm{s}$. When collisions are not avoided, these four models also reduce the contact speed only marginally, by less than 2\,km/h in all cases, whereas Alpamayo~1.5 and Alpamayo~2 reduce it by up to 7.25 and 4.22\,km/h, respectively.

On \gls{reactpai}, the pattern changes: at $\tau \geq 1.5\,\mathrm{s}$, both Alpamayo models achieve larger safety improvements for crossing pedestrians than for stationary cars. Inspecting the models' CoTs reveals that they describe the inserted car as stationary in only about a third of the cases for Alpamayo~1.5 (33\%) and a quarter for Alpamayo~2 (24\%). This suggests a possible training bias, with the models expecting leading vehicles to be in motion. When collisions are not avoided, both models also reduce the contact speed more than on \gls{reactnav}, by up to 10.25\,km/h for Alpamayo~1.5 and 8.49\,km/h for Alpamayo~2. For standing pedestrians, however, the two models differ markedly: the safety improvement $\dP$ of Alpamayo~2 is less than half that of Alpamayo~1.5 ($+15.89$ vs.\ $+35.69$ percentage points at $\tau = 2.0\,\mathrm{s}$), even though both recognize the pedestrian as standing in the lane at similar rates (60\% for Alpamayo~1.5 and 56\% for Alpamayo~2). The difference lies instead in how strongly they brake after recognizing it: over the final 0.5\,s of the horizon, Alpamayo~1.5 is on average 13.0\,km/h slower than in its original trajectory, compared with 7.7\,km/h for Alpamayo~2. For crossing pedestrians, both models perform similarly, likely because milder braking often suffices, as the pedestrian may clear the path before the vehicle arrives.

\begin{figure}[t]
    \centering
    \includegraphics[width=\linewidth]{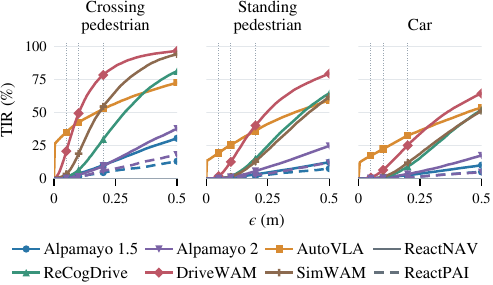}

    \caption{Trajectory invariance rate, $\TIR$: the percentage of samples whose original and edited trajectories differ by an \gls{ade} below $\epsilon$, i.e., for which the model essentially ignores the inserted object (higher means less reactive). Results are aggregated across all four look-ahead times $\tau$.}
    \label{fig:cum_ade}
\end{figure}

\Cref{fig:cum_ade} reports $\TIR$ for each inserted object at four thresholds, $\epsilon \in \{0.05, 0.1, 0.2, 0.5\}\,\mathrm{m}$. In many cases, the models barely react to the inserted objects at all. In the most extreme case, 34.9\% of AutoVLA's trajectory pairs have an \gls{ade} below 0.05\,m when a crossing pedestrian is inserted.

\subsection{Failure Analysis}
\label{subsec:ablations}
To understand where these failures originate, we study (a) whether the Alpamayo models reason correctly about the hazard and whether they act on this reasoning, and (b) how the individual training stages of ReCogDrive and SimWAM affect their reaction capabilities.

\paragraph{CoT Analysis.}
Of the evaluated models, only the two Alpamayo models consistently reason in natural language before predicting a trajectory. AutoVLA also supports reasoning, but in practice always defaults to its fast mode, producing the same generic reasoning output for every sample, so we exclude it from this analysis. We run all experiments in this section on \gls{reactpai}, which is closer to the training distribution of the Alpamayo models.

\begin{figure}[t]
    \centering
    \includegraphics[width=\linewidth]{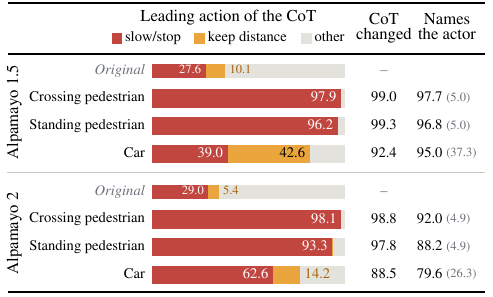}

    \caption{Effect of inserted objects on the CoTs of Alpamayo~1.5 and Alpamayo~2 on \gls{reactpai}, aggregated across all four look-ahead times $\tau$. Bars show the distribution of the CoT's leading action: \emph{slow/stop} (stop, yield, or slow down), \emph{keep distance} (keep a distance from or follow a lead vehicle), or \emph{other}. \emph{CoT changed}: percentage of edited-scene CoTs that differ from the original-scene CoT of the same sample. \emph{Names the actor}: percentage of edited-scene CoTs that mention a pedestrian or, for inserted cars, a vehicle (in parentheses: original-scene CoTs).}
    \label{fig:cot_summary}
\end{figure}

\Cref{fig:cot_summary} illustrates how the inserted objects affect the CoTs of the Alpamayo models. For the vast majority of samples with an inserted pedestrian, the CoT switches to a \emph{Slow/stop} instruction that explicitly refers to the inserted actor. The model thus detects the actor and correctly reasons that the vehicle should slow down or stop, yet in most cases the executed trajectory does not decelerate enough to avoid a collision. This suggests that, in this case, the main cause of failure in hazardous scenarios is not perception or reasoning, but rather a ``miscommunication'' between the \gls{vlm} backbone and the action expert, a phenomenon also reported in other recent studies~\cite{cercola2026perceives, mayumu2026vla, foutter2026vision, manalu2026measuring}. For inserted cars, the CoT of Alpamayo~1.5 most often changes to a \emph{Keep distance} instruction, consistent with the model mostly failing to recognize that the vehicle is stationary.

\begin{figure}[t]
    \centering
    \includegraphics[width=\linewidth]{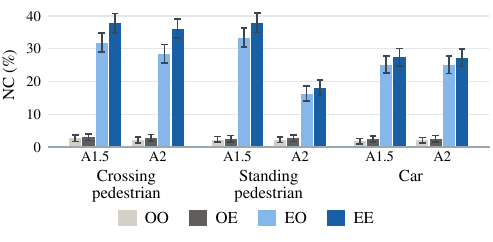}

    \caption{Effect of CoT swapping on $\NC$ for Alpamayo~1.5 (A1.5) and Alpamayo~2 (A2) on \gls{reactpai} at $\tau = 2.0\,\mathrm{s}$. The first and second letters give the source of the visual input and of the CoT: original (O) or edited (E). Whiskers indicate 95\% bootstrap CIs.}
    \label{fig:cot_swap}
\end{figure}

To investigate the causal effect of the CoT on the safety of the predicted trajectory, we swap the CoTs generated for the original and edited scenes, obtaining the trajectories $\mathbf{y}^{\mathrm{OE}} = \pi_\theta(I^{\mathrm{O}}, C^{\mathrm{E}})$ and $\mathbf{y}^{\mathrm{EO}} = \pi_\theta(I^{\mathrm{E}}, C^{\mathrm{O}})$, where $I$ denotes the visual input, $C$ the CoT, and the superscripts $\mathrm{O}$ and $\mathrm{E}$ the original and edited versions, respectively. As \cref{fig:cot_swap} shows for a look-ahead time of $\tau = 2.0\,\mathrm{s}$, the largest safety benefit appears to come from directly observing the inserted object ($I^{\mathrm{E}}$), rather than from receiving a CoT that instructs the model to yield to a pedestrian or car absent from its visual input. Once the object is visible, however, the CoT does help: the CoT generated for the edited scene leads to safer trajectories than the one generated for the original scene ($\mathbf{y}^{\mathrm{EE}}$ vs.\ $\mathbf{y}^{\mathrm{EO}}$). Even an explicit instruction to stop immediately, though, still leaves both models colliding in most samples (Appendix~\ref{subsec:supp_cot_analysis}).

\paragraph{Training Stages Analysis.}
ReCogDrive and SimWAM both have a released checkpoint for each training stage, which allows us to examine how each stage affects a model's ability to avoid imminent hazards. ReCogDrive has an additional initial stage in which its \gls{vlm} undergoes \gls{sft} on driving scenes across a range of tasks, including trajectory prediction. Apart from this stage, both models follow a similar recipe. They are first trained with imitation learning (IL) to match human expert trajectories and then fine-tuned with reinforcement learning (RL). The RL stage uses the \gls{pdms}~\cite{NEURIPS2024_32768f7f} as the reward and variants of GRPO~\cite{shao2024deepseekmath} adapted to diffusion (ReCogDrive) and flow-matching (SimWAM) models.

\Cref{tab:improv_by_training_stage} reports the safety improvement $\dP$ on \gls{reactnav} at $\tau = 2.0\,\mathrm{s}$ for each training-stage checkpoint of both models. For comparison with nominal driving performance, we also include each checkpoint's PDMS on navtest. We restrict the comparison to ego-vehicle speeds below 20\,km/h, since at higher speeds $\dP$ is close to zero for all checkpoints. For every inserted object, $\dP$ decreases from one training stage to the next, even though PDMS increases. The largest drop occurs between ReCogDrive's VLM SFT and IL stages, where $\dP$ for stationary cars falls from $+10.39$ to $+4.88$ (though the two stages also differ in their action head), whereas for crossing pedestrians, which all checkpoints rarely avoid, the differences are small. These results suggest that further optimizing for nominal driving performance, on data in which safety-critical scenarios are rare, can in fact degrade a model's ability to handle them.

\begin{table}[t]
    \centering
    \caption{Safety improvement $\dP$ (in percentage points) on \gls{reactnav} at $\tau = 2.0\,\mathrm{s}$ for each training-stage checkpoint of ReCogDrive and SimWAM, over samples with an initial ego-vehicle speed below 20\,km/h ($n = 1762$). Gray: the paired 95\% bootstrap CI of the difference to the previous training stage includes zero. PDMS on navtest as reported in~\cite{li2026recogdrive,zhao2026simwam}.}
    \label{tab:improv_by_training_stage}
    \footnotesize
    \setlength{\tabcolsep}{4pt}
    \begin{tabular}{@{}llcrrr@{}}
        \toprule
        & & & \multicolumn{3}{c}{$\dP$ ($v < 20\,\mathrm{km/h}$)} \\
        \cmidrule(l){4-6}
        Model & Stage & PDMS & Cross.\ ped. & Stand.\ ped. & Car \\
        \midrule
        \multirow{3}{*}{ReCogDrive} & VLM SFT & 84.1 & $+2.38$ & $+4.99$ & $+10.39$ \\
         & IL & 86.5 & $+0.85$ & $+1.36$ & $+4.88$ \\
         & RL & 90.8 & \textcolor{gray}{$+0.51$} & $+0.62$ & $+3.29$ \\
        \midrule
        \multirow{2}{*}{SimWAM} & IL & 90.3 & $+0.45$ & $+2.10$ & $+6.13$ \\
         & RL & 91.9 & \textcolor{gray}{$+0.40$} & $+1.25$ & $+4.14$ \\
        \bottomrule
    \end{tabular}
\end{table}

\section{Can Models Learn to React?}
\label{sec:teaching_to_react}

Having established that current \glspl{vla} and \glspl{wam} fail to handle safety-critical scenarios reliably, and in particular to brake in an emergency, we now ask how difficult it is to teach this behavior to an existing model, ReCogDrive, using edited samples similar to those in \benchname, without degrading its nominal driving performance and in a way that generalizes to scenes not seen during training. Since our hazardous edited scenes come without ground-truth trajectories, the most straightforward approach would be to add them to the RL training mix. However, GRPO can only reinforce behavior that the policy already samples: a group of candidate trajectories that all receive the same reward yields no gradient. We test this directly on ReCogDrive's IL checkpoint, i.e., the initialization of its RL stage. To this end, we create 300 hazardous scenes from navtrain, the training split of NAVSIM, by inserting the three \benchname objects exactly as in \gls{reactnav}. Even with a group size of $G = 32$, four times the one used in its RL stage, only 26.3\% of the groups yield a non-zero gradient, and only 1.3\% would reinforce a trajectory that is safe under \benchname's definition (Appendix~\ref{subsec:supp_rl_probe}). We therefore construct artificial target trajectories for the hazardous scenes and teach the desired behavior via supervised fine-tuning.

\subsection{SFT on Artificial Targets}
\label{subsec:sft_on_artificial_targets}

\paragraph{Training Data.}
Since \gls{reactnav} is drawn from navtest, we build our training data from navtrain: we randomly select 1,957 navtrain samples whose ego speed falls within the range covered by \gls{reactnav} and, for each of the three inserted objects, create edited variants at two of the four look-ahead times $\tau \in \{0.5, 1.0, 1.5, 2.0\}\,\mathrm{s}$, assigned such that all four are equally represented overall. This yields a total of 11,742 samples. To prevent the model from learning the shortcut that any visual artifact introduced by an inserted object is a cue to brake, we also create benign variants of the same samples, with the same objects and look-ahead times but with the objects placed off the ego vehicle's path, yielding another 11,742 benign samples. We refer to the resulting 23,484 samples as ReactTrain and provide further details and visualizations in Appendix~\ref{subsec:supp_reacttrain}.

\paragraph{Artificial Target Trajectories.}
Since the hazardous ReactTrain samples come without ground-truth trajectories that react to the inserted object, we construct artificial targets for supervised fine-tuning with a simple recipe. We first fit a cubic spline to the ground-truth waypoints, which preserves the original path while allowing us to change the speed profile along it. We then place the desired stopping point on the spline at a margin $d_{\mathrm{m}}$ before the inserted object and denote by $\Delta s$ the arc length between the ego vehicle and this point. Given the initial ego speed $v_0$, we use the constant deceleration $a = \min\{v_0^2 / (2\,\Delta s),\, a_{\max}\}$, i.e., the deceleration required to stop within $\Delta s$, capped at a maximum value $a_{\max}$. Finally, we obtain the new waypoints by sampling the spline at the arc length $s(t) = v_0 t - \tfrac{1}{2} a t^2$ that the ego vehicle travels up to each waypoint timestamp $t$, keeping the vehicle in place once it has stopped, i.e., for $t \geq v_0 / a$. We set $d_{\mathrm{m}} = 1\,\mathrm{m}$ and $a_{\max} = 8\,\mathrm{m/s^2}$, a common estimate of the maximum braking deceleration of a passenger car on a dry road~\cite{greibe2007braking,8813885}. We keep the samples in which even $a_{\max}$ cannot prevent the collision, all with $\tau = 0.5\,\mathrm{s}$, since their targets still teach the model to slow down as much as possible. We illustrate this construction in Appendix~\ref{subsec:supp_finetuning}.

\paragraph{Experimental Details and Results.}
\label{subsubsec:experimental_details_results}
We fine-tune the ReCogDrive IL checkpoint on ReactTrain with the denoising objective of its imitation-learning stage. To preserve nominal driving performance, each epoch additionally includes 11,742 original navtrain samples, drawn anew in every epoch, so that hazardous, benign, and original samples each make up one third of it. We supervise the hazardous samples with the artificial targets and the benign and original ones with the logged ground-truth trajectories. We update only the action head, keeping the \gls{vlm} frozen, and train for only three epochs, i.e., 276 gradient steps (Appendix~\ref{subsec:supp_finetuning}).

\Cref{tab:sft_results} compares the fine-tuned checkpoint (IL + RT) with the original IL checkpoint. The fine-tuned model reacts far more strongly to the inserted hazards on \gls{reactnav}, while its PDMS on navtest drops only marginally, mainly due to lower ego progress, while $\NC$ and TTC improve slightly. On the original \gls{reactnav} scenes, IL + RT plans at least 20\% less distance than IL in only 0.9\% of the samples, never braking harder than 4.5\,m/s$^2$.

\begin{table}[t]
    \centering
    \caption{Results of ReCogDrive's IL checkpoint (IL) and of the same checkpoint fine-tuned on ReactTrain (IL + RT) on \gls{reactnav}, aggregated across all look-ahead times $\tau$, and on navtest (PDMS and its sub-metrics on LQR-executed trajectories, from our own evaluation). Units are as in \cref{tab:main_results}.}
    \label{tab:sft_results}
    \footnotesize
    \setlength{\tabcolsep}{3pt}
    \begin{tabular}{@{}lcccccccc@{}}
        \toprule
        & \multicolumn{2}{c}{ReactNAV} & \multicolumn{6}{c}{navtest} \\
        \cmidrule(lr){2-3} \cmidrule(lr){4-9}
        & $\dP{\uparrow}$ & $\CSC{\downarrow}$ & $\NC{\uparrow}$ & $\DAC{\uparrow}$ & TTC$\uparrow$ & EP$\uparrow$ & C$\uparrow$ & PDMS$\uparrow$ \\
        \midrule
        IL  & $+0.5$        & $-0.3$          & 98.1          & 94.4          & 94.2          & \textbf{80.6} & \textbf{100.0} & \textbf{86.2} \\
        IL + RT & $\mathbf{+41.4}$ & $\mathbf{-9.9}$ & \textbf{98.7} & \textbf{94.5} & \textbf{94.9} & 79.3          & 99.9  & 86.0          \\
        \bottomrule
    \end{tabular}
\end{table}

\paragraph{Generalization.}
\label{subsubsec:generalization}
Finally, we examine whether the fine-tuned model has learned to slow down whenever its path is blocked, rather than merely reacting to the objects seen during training or to compositing artifacts. To this end, we construct two additional test sets from navtest (details in Appendix~\ref{subsec:supp_generalization_sets}). \emph{Unseen objects:} we insert standing animals, an object category absent from ReactTrain, into the ego vehicle's path in 291 samples at the same four look-ahead times. \emph{Real scenes:} we select 308 samples in which the nearest road user in the ego vehicle's path is a nearly stationary vehicle 5--15\,m ahead, and edit only the ego status, so that the ego vehicle appears to be traveling at a constant speed that would bring it into collision with this vehicle after the look-ahead time $\tau$. This produces the same kind of imminent hazard as in \gls{reactnav}, but with fully real visual input, so any reaction cannot be attributed to compositing artifacts. Since ReCogDrive sees only the current frame, the edited ego status does not contradict its visual input.

\begin{figure}[t]
    \centering
    \includegraphics[width=\linewidth]{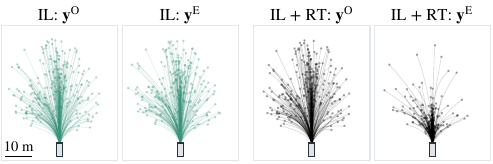}

    \caption{Plans of the IL and IL + RT checkpoints for the 291 unseen-objects samples, each at one random $\tau$, in the original ($\mathbf{y}^{\mathrm{O}}$) and edited ($\mathbf{y}^{\mathrm{E}}$) scenes; each line traces the front bumper over 4\,s. Although ReactTrain contains no animals, IL + RT shortens its plans in the edited scenes, while its original-scene plans stay close to those of the IL checkpoint.}
    \label{fig:recogdrive_rt_trajectories}
\end{figure}

\Cref{tab:generalization_results} reports the results for both test sets, aggregated across all four look-ahead times $\tau$. The fine-tuned model generalizes to both settings. On standing animals, the safety gains are smaller than on \gls{reactnav} ($+28.1$ vs.\ $+41.4$ in $\dP$), but still substantial relative to the IL checkpoint, which barely reacts (see also \cref{fig:recogdrive_rt_trajectories}). For the real scenes, which lack an original trajectory, we report absolute metrics; though not directly comparable to the deltas above, they show that the safety improvements carry over to fully real visual input ($54.2$ vs.\ $6.8$ in $P$), the setting that ultimately matters most.

\paragraph{Discussion.}
Overall, our results suggest that, in the models we probed, the generalization inherited from the foundation model often extends to recognizing hazards but not to acting on them. For inserted pedestrians, the Alpamayo CoTs call for a stop that the trajectory mostly does not execute (\cref{subsec:ablations}), and ReCogDrive brakes for unseen objects and in real scenes after only its action head is trained on the behavior, suggesting that its frozen \gls{vlm} already encodes whether the path ahead is blocked. This points to a distinction between long-tail scenarios, which are open-ended, and the behaviors these scenarios demand, such as emergency braking, which form a much smaller set~\cite{thorn2018framework}. Such behaviors are rare in nominal driving but, unlike the scenarios, can be covered explicitly, even with edited data such as ReactTrain, so that a model can draw on them when it faces a scenario it has never seen.

\begin{table}[t]
  \centering
  \caption{Generalization of the IL and IL + RT checkpoints to unseen objects and real scenes, aggregated across all four look-ahead times $\tau$. Real scenes have no original trajectory, so we report absolute values: $P$ in percent and the mean contact speed, $\CS$, in km/h.}
  \label{tab:generalization_results}
  \footnotesize
  \begin{tabular}{l cc cc}
    \toprule
    & \multicolumn{2}{c}{Unseen objects} & \multicolumn{2}{c}{Real scenes} \\
    \cmidrule(lr){2-3} \cmidrule(lr){4-5}
    & $\dP{\uparrow}$ & $\CSC{\downarrow}$
    & $P{\uparrow}$ & $\CS{\downarrow}$ \\
    \midrule
    IL   & $+0.1$ & $-0.2$ & $6.8$ & $21.4$ \\
    IL + RT & $\mathbf{+28.1}$ & $\mathbf{-8.7}$
         & $\mathbf{54.2}$ & $\mathbf{14.3}$ \\
    \bottomrule
  \end{tabular}
\end{table}

\section{Conclusion}
We introduced \benchname, a benchmark of exact, pixel-level counterfactual pairs built from real driving scenes, and used it to show that six recent \glspl{vla} and \glspl{wam} largely fail to handle safety-critical scenarios in which hard braking or a quick but safe nudge is needed to avoid a collision, calling into question the long-tail generalization often attributed to these models. At the same time, a short fine-tuning stage on edited scenes substantially improved how ReCogDrive reacts to imminent hazards, also for unseen objects and fully real scenes, at almost no cost to nominal driving. This suggests that models such as ReCogDrive can serve as a strong foundation for acquiring safety-critical behaviors such as emergency braking, and that training data could target these comparatively few behaviors directly, relying on the model to generalize them across the many long-tail scenarios in which they are needed.

\paragraph{Limitations.} 
\benchname covers mainly three object types and a single scenario (a close-range obstacle in the ego path), and its inserted objects are not fully photorealistic. Our evaluation is open-loop, so delayed reactions spanning several planning cycles are not assessed. Finally, our simple fine-tuning recipe leaves room for better uses of ReactTrain that may even avoid the small drop in nominal performance.

\section*{LLM Usage Statement}
We used two coding agents, Claude Code and Codex, in both the research and the manuscript. Under the authors' direction, the agents implemented, tested, and debugged parts of the code, and generated figures. The authors designed the benchmark, its metrics, the experiments, and the training recipe, reviewed all agent-generated figures, and validated the code and all reported numbers. The research questions, hypotheses, and conclusions are the authors' own; the agents were used only to critique and refine them. For the manuscript, the agents drafted parts of the appendix and helped find and summarize related work; the authors checked all generated text and verified every citation. The agents were also used to improve the language of the author-written main text, make notation consistent, and propose ways to shorten sections. The authors take full responsibility for the content of this paper.

{
    \small
    \bibliographystyle{ieeenat_fullname}
    \bibliography{main}

\begin{thebibliography}{57}
\providecommand{\natexlab}[1]{#1}
\providecommand{\url}[1]{\texttt{#1}}
\expandafter\ifx\csname urlstyle\endcsname\relax
  \providecommand{\doi}[1]{doi: #1}\else
  \providecommand{\doi}{doi: \begingroup \urlstyle{rm}\Url}\fi

\bibitem[Bao et~al.(2026)Bao, Ruan, Wang, Zhao, Leng, Zhang, Zeng, McAllister, Christensen, and Zhou]{bao2026navsafe}
Yuxin Bao, Hongwei Ruan, Luobin Wang, Seth~Z Zhao, Ziyang Leng, Zihan Zhang, Yu Zeng, Rowan McAllister, Henrik Christensen, and Bolei Zhou.
\newblock {NavSafe-$\infty$}: Benchmarking closed-loop driving safety in photorealistic environments.
\newblock \emph{arXiv preprint arXiv:2609.26618}, 2026.

\bibitem[{Blender Online Community}(2024)]{blender}
{Blender Online Community}.
\newblock Blender -- a {3D} modelling and rendering package.
\newblock \url{https://www.blender.org}, 2024.
\newblock Blender Foundation, Amsterdam. Version 4.1.

\bibitem[Caesar et~al.(2020)Caesar, Bankiti, Lang, Vora, Liong, Xu, Krishnan, Pan, Baldan, and Beijbom]{9156412}
Holger Caesar, Varun Bankiti, Alex~H. Lang, Sourabh Vora, Venice~Erin Liong, Qiang Xu, Anush Krishnan, Yu Pan, Giancarlo Baldan, and Oscar Beijbom.
\newblock {nuScenes}: A multimodal dataset for autonomous driving.
\newblock In \emph{2020 IEEE/CVF Conference on Computer Vision and Pattern Recognition (CVPR)}, pages 11618--11628, 2020.

\bibitem[Cao et~al.(2025)Cao, Hallgarten, Li, Dauner, Gu, Wang, Miron, Aiello, Li, Gilitschenski, Ivanovic, Pavone, Geiger, and Chitta]{cao2025pseudosimulation}
Wei Cao, Marcel Hallgarten, Tianyu Li, Daniel Dauner, Xunjiang Gu, Caojun Wang, Yakov Miron, Marco Aiello, Hongyang Li, Igor Gilitschenski, Boris Ivanovic, Marco Pavone, Andreas Geiger, and Kashyap Chitta.
\newblock Pseudo-simulation for autonomous driving.
\newblock In \emph{9th Annual Conference on Robot Learning}, 2025.

\bibitem[Cercola et~al.(2026)Cercola, Luo, Formentin, and Pavone]{cercola2026perceives}
Matteo Cercola, Katie~Z Luo, Simone Formentin, and Marco Pavone.
\newblock Perceives but doesn't brake: A perception-action binding gap in reasoning driving {VLAs}.
\newblock \emph{arXiv preprint arXiv:2610.28184}, 2026.

\bibitem[Dauner et~al.(2024)Dauner, Hallgarten, Li, Weng, Huang, Yang, Li, Gilitschenski, Ivanovic, Pavone, Geiger, and Chitta]{NEURIPS2024_32768f7f}
Daniel Dauner, Marcel Hallgarten, Tianyu Li, Xinshuo Weng, Zhiyu Huang, Zetong Yang, Hongyang Li, Igor Gilitschenski, Boris Ivanovic, Marco Pavone, Andreas Geiger, and Kashyap Chitta.
\newblock {NAVSIM}: Data-driven non-reactive autonomous vehicle simulation and benchmarking.
\newblock In \emph{Advances in Neural Information Processing Systems}, pages 28706--28719. Curran Associates, Inc., 2024.

\bibitem[Dosovitskiy et~al.(2017)Dosovitskiy, Ros, Codevilla, Lopez, and Koltun]{pmlr-v78-dosovitskiy17a}
Alexey Dosovitskiy, German Ros, Felipe Codevilla, Antonio Lopez, and Vladlen Koltun.
\newblock {CARLA}: {An} open urban driving simulator.
\newblock In \emph{Proceedings of the 1st Annual Conference on Robot Learning}, pages 1--16. PMLR, 2017.

\bibitem[Foutter et~al.(2026)Foutter, Cercola, Wild, Wang, Li, Gammelli, and Pavone]{foutter2026vision}
Matthew Foutter, Matteo Cercola, Lena Wild, Yunshan Wang, Michelle Li, Daniele Gammelli, and Marco Pavone.
\newblock Do vision-language-action models mean what they say? on the role of faithfulness in embodied reasoning.
\newblock \emph{arXiv preprint arXiv:2607.04681}, 2026.

\bibitem[Fu et~al.(2025)Fu, Zhang, Zhao, Cui, Liang, Zhang, Zhang, Xie, Wang, and Bai]{Fu_2025_ICCV}
Haoyu Fu, Diankun Zhang, Zongchuang Zhao, Jianfeng Cui, Dingkang Liang, Chong Zhang, Dingyuan Zhang, Hongwei Xie, Bing Wang, and Xiang Bai.
\newblock {ORION}: A holistic end-to-end autonomous driving framework by vision-language instructed action generation.
\newblock In \emph{Proceedings of the IEEE/CVF International Conference on Computer Vision (ICCV)}, pages 24823--24834, 2025.

\bibitem[Fu et~al.(2026)Fu, Yuan, Tian, Li, Zhu, Han, Zhang, Fang, Xue, Xu, et~al.]{fu20264d}
Jiacheng Fu, Yibo Yuan, Meng Tian, Yue Li, Jiangtong Zhu, Jianhua Han, Yueyi Zhang, Jianwu Fang, Jianru Xue, Hang Xu, et~al.
\newblock {4D-WAM}: {4D} consistent world modeling for autonomous driving.
\newblock \emph{arXiv preprint arXiv:2608.10107}, 2026.

\bibitem[Gassmann et~al.(2019)Gassmann, Oboril, Buerkle, Liu, Yan, Elli, Alvarez, Aerrabotu, Jaber, van Beek, Iyer, and Weast]{8813885}
Bernd Gassmann, Fabian Oboril, Cornelius Buerkle, Shuang Liu, Shoumeng Yan, Maria~Soledad Elli, Ignacio Alvarez, Naveen Aerrabotu, Suhel Jaber, Peter van Beek, Darshan Iyer, and Jack Weast.
\newblock Towards standardization of {AV} safety: C++ library for {Responsibility Sensitive Safety}.
\newblock In \emph{2019 IEEE Intelligent Vehicles Symposium (IV)}, pages 2265--2271, 2019.

\bibitem[Geng et~al.(2025)Geng, Du, Zhang, Li, Wang, and Huang]{geng2025driving}
Jiaheng Geng, Jiatong Du, Xinyu Zhang, Ye Li, Panqu Wang, and Yanjun Huang.
\newblock Driving in corner case: A real-world adversarial closed-loop evaluation platform for end-to-end autonomous driving.
\newblock \emph{arXiv preprint arXiv:2512.16055}, 2025.

\bibitem[Gerstenecker et~al.(2026)Gerstenecker, Geiger, and Renz]{gerstenecker2026fail2drive}
Simon Gerstenecker, Andreas Geiger, and Katrin Renz.
\newblock {Fail2Drive}: Benchmarking closed-loop driving generalization.
\newblock \emph{arXiv preprint arXiv:2604.08535}, 2026.

\bibitem[Gonzalez-Franco et~al.(2020)Gonzalez-Franco, Ofek, Pan, Antley, Steed, Spanlang, Maselli, Banakou, Pelechano, Orts-Escolano, Orvalho, Trutoiu, Wojcik, Sanchez-Vives, Bailenson, Slater, and Lanier]{10.3389/frvir.2020.561558}
Mar Gonzalez-Franco, Eyal Ofek, Ye Pan, Angus Antley, Anthony Steed, Bernhard Spanlang, Antonella Maselli, Domna Banakou, Nuria Pelechano, Sergio Orts-Escolano, Veronica Orvalho, Laura Trutoiu, Markus Wojcik, Maria~V. Sanchez-Vives, Jeremy Bailenson, Mel Slater, and Jaron Lanier.
\newblock The {Rocketbox} library and the utility of freely available rigged avatars.
\newblock \emph{Frontiers in Virtual Reality}, Volume 1 - 2020, 2020.

\bibitem[Greibe(2007)]{greibe2007braking}
Poul Greibe.
\newblock Braking distance, friction and behaviour: Findings, analyses and recommendations based on braking trials.
\newblock Technical report, Trafitec, 2007.

\bibitem[Guerreiro et~al.(2023)Guerreiro, Nakazawa, and Stenger]{10204242}
Julian Jorge~Andrade Guerreiro, Mitsuru Nakazawa, and Björn Stenger.
\newblock {PCT-Net}: Full resolution image harmonization using pixel-wise color transformations.
\newblock In \emph{2023 IEEE/CVF Conference on Computer Vision and Pattern Recognition (CVPR)}, pages 5917--5926, 2023.

\bibitem[Hallgarten et~al.(2024)Hallgarten, Zapata, Stoll, Renz, and Zell]{10803052}
Marcel Hallgarten, Julian Zapata, Martin Stoll, Katrin Renz, and Andreas Zell.
\newblock Can vehicle motion planning generalize to realistic long-tail scenarios?
\newblock In \emph{2024 IEEE/RSJ International Conference on Intelligent Robots and Systems (IROS)}, pages 5388--5395, 2024.

\bibitem[Hwang et~al.(2025)Hwang, Xu, Lin, Hung, Ji, Choi, Huang, He, Covington, Sapp, Zhou, Guo, Anguelov, and Tan]{hwang2025emma}
Jyh-Jing Hwang, Runsheng Xu, Hubert Lin, Wei-Chih Hung, Jingwei Ji, Kristy Choi, Di Huang, Tong He, Paul Covington, Benjamin Sapp, Yin Zhou, James Guo, Dragomir Anguelov, and Mingxing Tan.
\newblock {EMMA}: End-to-end multimodal model for autonomous driving.
\newblock \emph{Transactions on Machine Learning Research}, 2025.

\bibitem[Jia et~al.(2024)Jia, Yang, Li, Zhang, and Yan]{NEURIPS2024_017761f9}
Xiaosong Jia, Zhenjie Yang, Qifeng Li, Zhiyuan Zhang, and Junchi Yan.
\newblock {Bench2Drive}: Towards multi-ability benchmarking of closed-loop end-to-end autonomous driving.
\newblock In \emph{Advances in Neural Information Processing Systems}, pages 819--844. Curran Associates, Inc., 2024.

\bibitem[Jiang et~al.(2025)Jiang, Song, Li, Zhu, and Zhang]{jiang2025realengine}
Junzhe Jiang, Nan Song, Jingyu Li, Xiatian Zhu, and Li Zhang.
\newblock {RealEngine}: Simulating autonomous driving in realistic context.
\newblock \emph{arXiv preprint arXiv:2505.16902}, 2025.

\bibitem[Kerbl et~al.(2023)Kerbl, Kopanas, Leimk{\"u}hler, and Drettakis]{kerbl:hal-04088161}
Bernhard Kerbl, Georgios Kopanas, Thomas Leimk{\"u}hler, and George Drettakis.
\newblock {3D Gaussian Splatting for Real-Time Radiance Field Rendering}.
\newblock \emph{{ACM Transactions on Graphics}}, 42\penalty0 (4):\penalty0 1--14, 2023.

\bibitem[Li et~al.(2026)Li, Xiong, Guo, Li, Yan, Xu, Zhou, Chen, Sun, Wang, Ma, Chen, Ye, Liu, and Wang]{li2026recogdrive}
Yongkang Li, Kaixin Xiong, Xiangyu Guo, Fang Li, Sixu Yan, Gangwei Xu, Lijun Zhou, Long Chen, Haiyang Sun, Bing Wang, Kun Ma, Guang Chen, Hangjun Ye, Wenyu Liu, and Xinggang Wang.
\newblock {ReCogDrive}: A reinforced cognitive framework for end-to-end autonomous driving.
\newblock In \emph{The Fourteenth International Conference on Learning Representations}, 2026.

\bibitem[Liu et~al.(2026)Liu, Zhang, Liu, Cui, Xie, Chen, Ye, Nex, Cheng, and Yang]{liu2026universe}
Mengmeng Liu, Diankun Zhang, Jiuming Liu, Jianfeng Cui, Hongwei Xie, Guang Chen, Hangjun Ye, Francesco Nex, Hao Cheng, and Michael~Ying Yang.
\newblock {UNIVERSE}: Unified video action models for autonomous driving with flexible mask-modulated modality generation.
\newblock \emph{arXiv preprint arXiv:2607.05133}, 2026.

\bibitem[Ljungbergh et~al.(2025)Ljungbergh, Tonderski, Johnander, Caesar, {\AA}str{\"o}m, Felsberg, and Petersson]{10.1007/978-3-031-73404-5_10}
William Ljungbergh, Adam Tonderski, Joakim Johnander, Holger Caesar, Kalle {\AA}str{\"o}m, Michael Felsberg, and Christoffer Petersson.
\newblock {NeuroNCAP}: Photorealistic closed-loop safety testing for autonomous driving.
\newblock In \emph{Computer Vision -- ECCV 2024}, pages 161--177, Cham, 2025. Springer Nature Switzerland.

\bibitem[Luo et~al.(2026)Luo, Sun, Lu, Zhou, Wu, Li, Qin, and Li]{luo2026exceptiondrive}
Ziyi Luo, Zhe Sun, Yehao Lu, Lei Zhou, Lisheng Wu, Xuewei Li, Zequn Qin, and Xi Li.
\newblock {ExceptionDrive}: A planning-oriented counterfactual corner-case benchmark for autonomous driving.
\newblock \emph{arXiv preprint arXiv:2609.37871}, 2026.

\bibitem[Manalu et~al.(2026)Manalu, Wazir, Do, Kim, Hwang, Kidane, and Kim]{manalu2026measuring}
Gabriel Manalu, Saad Wazir, Thao Do, Seongah Kim, Hyeontaek Hwang, Tolcha~Yalew Kidane, and Daeyoung Kim.
\newblock Measuring the reasoning-action coupling in driving vision-language-action models through counterfactuals.
\newblock \emph{arXiv preprint arXiv:2610.30953}, 2026.

\bibitem[Mayumu et~al.(2026)Mayumu, Deng, and Mukala]{mayumu2026vla}
Nicanor Mayumu, Xiaoheng Deng, and Patrick Mukala.
\newblock Is {VLA} reasoning faithful? {Probing} safety of chain-of-causation in autonomous driving models.
\newblock \emph{arXiv preprint arXiv:2605.17268}, 2026.

\bibitem[Mildenhall et~al.(2021)Mildenhall, Srinivasan, Tancik, Barron, Ramamoorthi, and Ng]{10.1145/3503250}
Ben Mildenhall, Pratul~P. Srinivasan, Matthew Tancik, Jonathan~T. Barron, Ravi Ramamoorthi, and Ren Ng.
\newblock {NeRF}: representing scenes as neural radiance fields for view synthesis.
\newblock \emph{Commun. ACM}, 65\penalty0 (1):\penalty0 99–106, 2021.

\bibitem[Niu et~al.(2021)Niu, Cong, Liu, Hong, Zhang, Liang, and Zhang]{niu2021making}
Li Niu, Wenyan Cong, Liu Liu, Yan Hong, Bo Zhang, Jing Liang, and Liqing Zhang.
\newblock Making images real again: A comprehensive survey on deep image composition.
\newblock \emph{arXiv preprint arXiv:2106.14490}, 2021.

\bibitem[{NVIDIA}(2025)]{nvidia2025paiav}
{NVIDIA}.
\newblock {PhysicalAI-Autonomous-Vehicles}.
\newblock \url{https://huggingface.co/datasets/nvidia/PhysicalAI-Autonomous-Vehicles}, 2025.
\newblock Dataset. Accessed: 2026-10-21.

\bibitem[{NVIDIA}(2026{\natexlab{a}})]{nvidia2026alpamayo15}
{NVIDIA}.
\newblock Alpamayo 1.5.
\newblock \url{https://huggingface.co/nvidia/Alpamayo-1.5-10B}, 2026{\natexlab{a}}.
\newblock Model card. Accessed: 2026-10-15.

\bibitem[{NVIDIA}(2026{\natexlab{b}})]{nvidia2026alpamayo2}
{NVIDIA}.
\newblock {Alpamayo 2 Super}.
\newblock \url{https://huggingface.co/nvidia/Alpamayo2-Super}, 2026{\natexlab{b}}.
\newblock Model card. Accessed: 2026-10-15.

\bibitem[Pai et~al.(2026)Pai, Achenbach, Sanchez, Charalambous, Montesinos, Forrai, Mees, and Nava]{PaiJ-RSS-26}
Jonas Pai, Liam Achenbach, Oliver Sanchez, Stefanos Charalambous, Victoriano Montesinos, Benedek Forrai, Oier Mees, and Elvis Nava.
\newblock {mimic-video: Video-Action Models for Generalizable Robot Control Beyond VLAs}.
\newblock In \emph{Proceedings of Robotics: Science and Systems}, Sydney, Australia, 2026.

\bibitem[Peng et~al.(2025)Peng, Li, Wu, Zhang, Sun, and Wu]{ICLR2025_6cf58a87}
Yansong Peng, Hebei Li, Peixi Wu, Yueyi Zhang, Xiaoyan Sun, and Feng Wu.
\newblock {D-FINE}: Redefine regression task of {DETRs} as fine-grained distribution refinement.
\newblock In \emph{International Conference on Learning Representations}, pages 44015--44031, 2025.

\bibitem[Renz et~al.(2025)Renz, Chen, Arani, and Sinavski]{Renz_2025_CVPR}
Katrin Renz, Long Chen, Elahe Arani, and Oleg Sinavski.
\newblock {SimLingo}: Vision-only closed-loop autonomous driving with language-action alignment.
\newblock In \emph{Proceedings of the IEEE/CVF Conference on Computer Vision and Pattern Recognition (CVPR)}, pages 11993--12003, 2025.

\bibitem[Shao et~al.(2024{\natexlab{a}})Shao, Hu, Wang, Song, Waslander, Liu, and Li]{Shao_2024_CVPR}
Hao Shao, Yuxuan Hu, Letian Wang, Guanglu Song, Steven~L. Waslander, Yu Liu, and Hongsheng Li.
\newblock {LMDrive}: Closed-loop end-to-end driving with large language models.
\newblock In \emph{Proceedings of the IEEE/CVF Conference on Computer Vision and Pattern Recognition (CVPR)}, pages 15120--15130, 2024{\natexlab{a}}.

\bibitem[Shao et~al.(2024{\natexlab{b}})Shao, Wang, Zhu, Xu, Song, Bi, Zhang, Zhang, Li, et~al.]{shao2024deepseekmath}
Zhihong Shao, Peiyi Wang, Qihao Zhu, Runxin Xu, Junxiao Song, Xiao Bi, Haowei Zhang, Mingchuan Zhang, YK Li, et~al.
\newblock {DeepSeekMath}: Pushing the limits of mathematical reasoning in open language models.
\newblock \emph{arXiv preprint arXiv:2402.03300}, 2024{\natexlab{b}}.

\bibitem[Shi et~al.(2026)Shi, Xu, Shi, Sheng, Zhang, and Jiang]{shi2026drivewam}
Chen Shi, Jinrui Xu, Shaoshuai Shi, Kehua Sheng, Bo Zhang, and Li Jiang.
\newblock {DriveWAM}: Video generative priors enable scalable world-action modeling for autonomous driving.
\newblock \emph{arXiv preprint arXiv:2605.28544}, 2026.

\bibitem[Thorn et~al.(2018)Thorn, Kimmel, and Chaka]{thorn2018framework}
Eric Thorn, Shawn Kimmel, and Michelle Chaka.
\newblock A framework for automated driving system testable cases and scenarios.
\newblock Technical Report DOT HS 812 623, National Highway Traffic Safety Administration, 2018.

\bibitem[Tian et~al.(2024)Tian, Gu, Li, Liu, Wang, Zhao, Zhan, Jia, Lang, and Zhao]{tian2024drivevlm}
Xiaoyu Tian, Junru Gu, Bailin Li, Yicheng Liu, Yang Wang, Zhiyong Zhao, Kun Zhan, Peng Jia, XianPeng Lang, and Hang Zhao.
\newblock Drive{VLM}: The convergence of autonomous driving and large vision-language models.
\newblock In \emph{8th Annual Conference on Robot Learning}, 2024.

\bibitem[Tonderski et~al.(2024)Tonderski, Lindström, Hess, Ljungbergh, Svensson, and Petersson]{10657387}
Adam Tonderski, Carl Lindström, Georg Hess, William Ljungbergh, Lennart Svensson, and Christoffer Petersson.
\newblock {NeuRAD}: Neural rendering for autonomous driving.
\newblock In \emph{2024 IEEE/CVF Conference on Computer Vision and Pattern Recognition (CVPR)}, pages 14895--14904, 2024.

\bibitem[{vevalabs}(2025)]{vevalabs2025openxassets}
{vevalabs}.
\newblock {OpenX Assets}: Vehicle and traffic simulation assets using {ASAM OpenX} standards.
\newblock \url{https://github.com/vevalabs/openx-assets}, 2025.
\newblock Accessed: 2026-10-21.

\bibitem[Wang et~al.(2025{\natexlab{a}})Wang, Yu, Jiang, Lan, Shi, Chang, Kautz, Li, and Alvarez]{wang2025omnidrive}
Shihao Wang, Zhiding Yu, Xiaohui Jiang, Shiyi Lan, Min Shi, Nadine Chang, Jan Kautz, Ying Li, and Jose~M Alvarez.
\newblock {OmniDrive}: A holistic vision-language dataset for autonomous driving with counterfactual reasoning.
\newblock In \emph{2025 IEEE/CVF Conference on Computer Vision and Pattern Recognition (CVPR)}, pages 22442--22452. IEEE, 2025{\natexlab{a}}.

\bibitem[Wang et~al.(2025{\natexlab{b}})Wang, Luo, Bai, Cao, Che, Chen, Chen, Diamond, Ding, Ding, et~al.]{wang2025alpamayo}
Yan Wang, Wenjie Luo, Junjie Bai, Yulong Cao, Tong Che, Ke Chen, Yuxiao Chen, Jenna Diamond, Yifan Ding, Wenhao Ding, et~al.
\newblock {Alpamayo-R1}: Bridging reasoning and action prediction for generalizable autonomous driving in the long tail.
\newblock \emph{arXiv preprint arXiv:2511.00088}, 2025{\natexlab{b}}.

\bibitem[Xia et~al.(2026)Xia, Li, Zhou, Yao, Xiong, Sun, Wang, Ma, Chen, Ye, Liu, and Wang]{Xia_2026_CVPR}
Tianze Xia, Yongkang Li, Lijun Zhou, Jingfeng Yao, Kaixin Xiong, Haiyang Sun, Bing Wang, Kun Ma, Guang Chen, Hangjun Ye, Wenyu Liu, and Xinggang Wang.
\newblock {DriveLaW}: Unifying planning and video generation in a latent driving world.
\newblock In \emph{Proceedings of the IEEE/CVF Conference on Computer Vision and Pattern Recognition (CVPR)}, pages 39701--39712, 2026.

\bibitem[Xu et~al.(2022)Xu, Ding, Lyu, Liu, Wang, He, Hu, Zhao, and Li]{xu2022safebench}
Chejian Xu, Wenhao Ding, Weijie Lyu, Zuxin Liu, Shuai Wang, Yihan He, Hanjiang Hu, Ding Zhao, and Bo Li.
\newblock {SafeBench}: A benchmarking platform for safety evaluation of autonomous vehicles.
\newblock In \emph{Thirty-sixth Conference on Neural Information Processing Systems Datasets and Benchmarks Track}, 2022.

\bibitem[Xu et~al.(2026{\natexlab{a}})Xu, Lin, Jeon, Feng, Zou, Sun, Gorman, Tolstaya, Tang, White, Sapp, Tan, Hwang, and Anguelov]{Xu_2026_CVPR}
Runsheng Xu, Hubert Lin, Wonseok Jeon, Hao Feng, Yuliang Zou, Liting Sun, John Gorman, Kate Tolstaya, Sarah Tang, Brandyn White, Ben Sapp, Mingxing Tan, Jyh-Jing Hwang, and Dragomir Anguelov.
\newblock {WOD-E2E}: {Waymo Open Dataset} for end-to-end driving in challenging long-tail scenarios.
\newblock In \emph{Proceedings of the IEEE/CVF Conference on Computer Vision and Pattern Recognition (CVPR)}, pages 3709--3718, 2026{\natexlab{a}}.

\bibitem[Xu et~al.(2024)Xu, Zhang, Xie, Zhao, Guo, Wong, Li, and Zhao]{xu2024drivegpt4}
Zhenhua Xu, Yujia Zhang, Enze Xie, Zhen Zhao, Yong Guo, Kwan-Yee~K Wong, Zhenguo Li, and Hengshuang Zhao.
\newblock {DriveGPT4}: Interpretable end-to-end autonomous driving via large language model.
\newblock \emph{IEEE Robotics and Automation Letters}, 9\penalty0 (10):\penalty0 8186--8193, 2024.

\bibitem[Xu et~al.(2026{\natexlab{b}})Xu, Li, Gao, Gao, Jin, Zhao, Feng, Zhang, and Zhao]{xu2025challenger}
Zhiyuan Xu, Bohan Li, Huan-ang Gao, Mingju Gao, Xin Jin, Hang Zhao, Shuo Feng, Ya-Qin Zhang, and Hao Zhao.
\newblock Challenger: Affordable adversarial driving video generation for safety testing.
\newblock \emph{Proceedings of the IASEAI Conference}, 2\penalty0 (1):\penalty0 767--780, 2026{\natexlab{b}}.

\bibitem[Yang et~al.(2024)Yang, Kang, Huang, Zhao, Xu, Feng, and Zhao]{NEURIPS2024_26cfdcd8}
Lihe Yang, Bingyi Kang, Zilong Huang, Zhen Zhao, Xiaogang Xu, Jiashi Feng, and Hengshuang Zhao.
\newblock {Depth Anything V2}.
\newblock In \emph{Advances in Neural Information Processing Systems}, pages 21875--21911. Curran Associates, Inc., 2024.

\bibitem[Yang et~al.(2026)Yang, Huang, Wang, Wan, Luo, Sheng, and Chen]{yang2026beyond}
Ruolin Yang, Zilin Huang, Buoyue Wang, Zhengyang Wan, Yuhao Luo, Zihao Sheng, and Sikai Chen.
\newblock Beyond the leaderboard: Counterfactual diagnosis of end-to-end and {VLA} driving policies under domain shift.
\newblock \emph{arXiv preprint arXiv:2609.22582}, 2026.

\bibitem[Ye et~al.(2026)Ye, Ge, Zheng, Gao, Yu, Kurian, Indupuru, Tan, Zhu, Xiang, et~al.]{ye2026world}
Seonghyeon Ye, Yunhao Ge, Kaiyuan Zheng, Shenyuan Gao, Sihyun Yu, George Kurian, Suneel Indupuru, You~Liang Tan, Chuning Zhu, Jiannan Xiang, et~al.
\newblock World action models are zero-shot policies.
\newblock \emph{arXiv preprint arXiv:2602.15922}, 2026.

\bibitem[Zhao et~al.(2026)Zhao, Zhou, Xu, Sun, Zhou, Wu, Li, Liang, and Bai]{zhao2026simwam}
Zongchuang Zhao, Xin Zhou, Tianyang Xu, Zhengyang Sun, Kaixuan Zhou, Yu Wu, Honglin Li, Dingkang Liang, and Xiang Bai.
\newblock {SimWAM}: A simple world action model for end-to-end autonomous driving.
\newblock \emph{arXiv preprint arXiv:2608.07468}, 2026.

\bibitem[Zhou et~al.(2026{\natexlab{a}})Zhou, Lin, Wang, Lu, Bai, Liu, Wang, Geiger, and Liao]{11314796}
Hongyu Zhou, Longzhong Lin, Jiabao Wang, Yichong Lu, Dongfeng Bai, Bingbing Liu, Yue Wang, Andreas Geiger, and Yiyi Liao.
\newblock {HUGSIM}: A real-time, photo-realistic and closed-loop simulator for autonomous driving.
\newblock \emph{IEEE Transactions on Pattern Analysis and Machine Intelligence}, 48\penalty0 (4):\penalty0 4673--4691, 2026{\natexlab{a}}.

\bibitem[Zhou et~al.(2026{\natexlab{b}})Zhou, Lyu, Tian, Zhuo, and Li]{pmlr-v306-zhou26aj}
Jiawei Zhou, Linye Lyu, Zhuotao Tian, Cheng Zhuo, and Yu Li.
\newblock {SMD}: Multi-view safety-critical driving video generation in the real-world domain.
\newblock In \emph{Proceedings of the 43rd International Conference on Machine Learning}, pages 165568--165590. PMLR, 2026{\natexlab{b}}.

\bibitem[Zhou et~al.(2025)Zhou, Cai, Zhao, Zhang, Huang, Zhou, and Ma]{zhou2025autovla}
Zewei Zhou, Tianhui Cai, Seth~Z. Zhao, Yun Zhang, Zhiyu Huang, Bolei Zhou, and Jiaqi Ma.
\newblock Auto{VLA}: A vision-language-action model for end-to-end autonomous driving with adaptive reasoning and reinforcement fine-tuning.
\newblock In \emph{The Thirty-ninth Annual Conference on Neural Information Processing Systems}, 2025.

\bibitem[Zitkovich et~al.(2023)Zitkovich, Yu, Xu, Xu, Xiao, Xia, Wu, Wohlhart, Welker, Wahid, Vuong, Vanhoucke, Tran, Soricut, Singh, Singh, Sermanet, Sanketi, Salazar, Ryoo, Reymann, Rao, Pertsch, Mordatch, Michalewski, Lu, Levine, Lee, Lee, Leal, Kuang, Kalashnikov, Julian, Joshi, Irpan, Ichter, Hsu, Herzog, Hausman, Gopalakrishnan, Fu, Florence, Finn, Dubey, Driess, Ding, Choromanski, Chen, Chebotar, Carbajal, Brown, Brohan, Arenas, and Han]{pmlr-v229-zitkovich23a}
Brianna Zitkovich, Tianhe Yu, Sichun Xu, Peng Xu, Ted Xiao, Fei Xia, Jialin Wu, Paul Wohlhart, Stefan Welker, Ayzaan Wahid, Quan Vuong, Vincent Vanhoucke, Huong Tran, Radu Soricut, Anikait Singh, Jaspiar Singh, Pierre Sermanet, Pannag~R. Sanketi, Grecia Salazar, Michael~S. Ryoo, Krista Reymann, Kanishka Rao, Karl Pertsch, Igor Mordatch, Henryk Michalewski, Yao Lu, Sergey Levine, Lisa Lee, Tsang-Wei~Edward Lee, Isabel Leal, Yuheng Kuang, Dmitry Kalashnikov, Ryan Julian, Nikhil~J. Joshi, Alex Irpan, Brian Ichter, Jasmine Hsu, Alexander Herzog, Karol Hausman, Keerthana Gopalakrishnan, Chuyuan Fu, Pete Florence, Chelsea Finn, Kumar~Avinava Dubey, Danny Driess, Tianli Ding, Krzysztof~Marcin Choromanski, Xi Chen, Yevgen Chebotar, Justice Carbajal, Noah Brown, Anthony Brohan, Montserrat~Gonzalez Arenas, and Kehang Han.
\newblock {RT-2}: Vision-language-action models transfer web knowledge to robotic control.
\newblock In \emph{Proceedings of The 7th Conference on Robot Learning}, pages 2165--2183. PMLR, 2023.

\end{thebibliography}
}

% Keep supplementary-only citations out of the main reference list: from here on,
% \cite writes its \citation (and backref) entries to \jobname.scit instead of the .aux file.
% \makeatletter
% \newwrite\supp@citeout
% \immediate\openout\supp@citeout=\jobname.scit
% \let\supp@origcite\cite
% \let\supp@origbackout\Hy@backout
% \protected\def\cite#1{{%
%   \let\@auxout\supp@citeout
%   \def\@extra@b@citeb{@appx}%
%   \def\Hy@backout##1{{\let\@auxout\@mainaux\supp@origbackout{##1}}}%
%   \supp@origcite{#1}}}
% \makeatother
\clearpage
\setcounter{page}{1}
\maketitlesupplementary
\appendix

\section{Additional \benchname Details}
\label{sec:supp_benchmark}
\suppressfloats[t]

\subsection{Object Compositing}
\label{subsec:supp_object_compositing}
This section complements \cref{subsec:object_compositing} with implementation details of the compositing pipeline. \Cref{alg:asset_compositing} summarizes the pipeline, and the paragraphs below follow the order of its steps.

\begin{algorithm}[t]
    \caption{Compositing 3D assets into driving scenes}
    \label{alg:asset_compositing}
    \begin{algorithmic}[1]
        \Require Camera images $\{I_{t,c}\}$, camera calibration,
        LiDAR point clouds, ground-truth ego trajectory, 3D asset $A$,
        look-ahead time $\tau$
        \Ensure Edited images $\{\widetilde{I}_{t,c}\}$
        \State Select an insertion location using the ego vehicle's front-bumper position along its ground-truth trajectory at look-ahead time $\tau$
        \State Define the asset's motion or stationary placement
        \State Estimate the local ground surface from LiDAR
        \For{each history timestep $t$}
            \State Set the asset's 3D position, orientation, and pose on the estimated ground
            \For{each camera $c$}
                \State Render asset and contact shadow using camera calibration
                \State Mask occluded asset regions using scene depth
                and LiDAR evidence
                \State Blend the visible asset and shadow into $I_{t,c}$
                \State Harmonize the inserted foreground with PCT-Net
                to obtain $\widetilde{I}_{t,c}$
            \EndFor
        \EndFor
        \State \Return $\{\widetilde{I}_{t,c}\}$
    \end{algorithmic}
\end{algorithm}

\paragraph{Asset Selection and Placement.}
We use 42 human assets from the Microsoft Rocketbox library~\cite{10.3389/frvir.2020.561558} and seven vehicle assets from OpenX Assets~\cite{vevalabs2025openxassets}. Assets are assigned pseudo-randomly from the scene identifier with a fixed seed, so all four look-ahead variants of a scene use the same asset. Crossing pedestrians walk at 1.4\,m/s; consecutive history frames show successive phases of the asset's walk cycle and are rendered with motion blur. The crossing direction is also chosen pseudo-randomly per scene; if the placement is invalid in this direction at any look-ahead time, the opposite direction is used, so that all four variants of a scene share the same direction. Standing pedestrians use a natural two-foot standing pose and face the approaching ego vehicle. The stationary car is aligned with the ego vehicle's heading at the look-ahead time $\tau$, with its rear bumper placed exactly at the ego vehicle's front bumper at $\tau$.

\paragraph{Placement Validity.}
A placement is considered valid if (a) the object's bounding box (see Appendix~\ref{subsec:supp_metrics_trajectory}) contains fewer than five LiDAR returns higher than 0.15\,m above its base at each history position and, for crossing pedestrians, also at intervals of 0.25\,m along the remaining path to the interception point; (b) the ground fit described below is reliable in every history frame; (c) for the car, the ego vehicle's footprint does not touch it at any ground-truth pose before $\tau$, so that the first contact occurs exactly at $\tau$; and (d) the object is visible in at least one camera view.

\paragraph{Ground Estimation.}
In each history frame, we fit a plane to the LiDAR returns within $\pm 4$\,m of the object, using the lowest returns of each cell of a coarse grid to suppress other objects, followed by iterative outlier rejection. The fit is considered reliable if at least 80 returns lie within 12\,cm of the plane, at least 12 of them within 1.6\,m of the object, the root-mean-square residual of these returns is at most 9\,cm, and the gradient of the plane is at most 0.22. On \gls{reactnav}, pedestrians are grounded independently in each history frame, which absorbs small registration errors between frames, whereas the car is grounded once, in the last history frame, and keeps the same rigid pose in all frames, since independent fits would make it visibly jitter. On \gls{reactpai}, whose history frames are only 0.1\,s apart, all objects use a single plane: the median of the per-frame fits, expressed in a common frame. In both cases, the object is placed on the fitted plane, and its vertical axis is tilted from gravity towards the plane normal by at most $2^\circ$, so pedestrians stand upright, as on real slopes, and a noisy fit cannot visibly lean the object. 

\paragraph{Rendering and Lighting.}
Objects are rendered with an ideal pinhole camera and then warped into each camera's lens model: the OpenCV distortion model for \gls{reactnav} and an f-theta fisheye model for \gls{reactpai}. On \gls{reactpai}, we additionally motion-compensate the LiDAR returns using their individual timestamps and account for the rolling shutter when projecting them into the image. The object is lit only by an environment light whose colour is estimated from the sky regions of the camera images (or from their upper part if no sky is detected) and whose intensity is set from the median image brightness; we do not render direct sunlight or directional cast shadows. The contact shadow is obtained by rendering the object above a shadow-catcher plane aligned with the fitted ground under the same environment light.

\paragraph{Occlusion.}
Depth Anything V2 (Small)~\cite{NEURIPS2024_26cfdcd8} predicts relative inverse depth, which is known only up to an unknown scale and shift. We recover metric depth by projecting the LiDAR points into the image and robustly fitting this scale and shift to their inverse depths, which yields a dense metric depth map with the sharp boundaries of the monocular estimate. An object pixel with rendered depth $d$ is occluded if the scene depth at that pixel is smaller than $d - \max(0.55\,\mathrm{m},\, 0.045\,d)$; the margin, which grows with distance like the error of the monocular depth, prevents the background or the supporting surface from occluding the object. Pixels whose scene depth is consistent with the fitted ground plane are never treated as occluders, which protects the feet and wheels from depth errors near the road. Because monocular depth can be inaccurate, occlusions additionally require LiDAR evidence of a closer object: on \gls{reactnav}, a predicted near-total occlusion without LiDAR support is discarded, and on \gls{reactpai}, only connected occluding regions that contain LiDAR points in front of the object are kept. If the monocular depth cannot be aligned to the LiDAR, we skip occlusion handling for that view on \gls{reactpai}, whereas on \gls{reactnav} we discard the sample. Since occlusion handling can therefore miss occluders, we manually checked 100 randomly selected \gls{reactpai} views in which at least 500 pixels of the object are visible. In none of them is the object drawn over a closer real object that should hide it, or hidden where nothing real is in front of it. Hence, with 95\% confidence, fewer than 3\% of these views contain such an occlusion error. Finally, pixels showing the ego vehicle's body are never modified. On \gls{reactnav}, the ego vehicle is not visible in the front camera and only appears at the bottom edge of the front-left and front-right cameras (CAM\_L0/R0). On \gls{reactpai}, \gls{paiav} provides a mask of these pixels for each camera, which we use for all cameras except the front-wide one. For that camera, the provided mask is overly conservative, so we instead obtain it by projecting the hood of a 3D model of the ego vehicle (Mercedes-Benz GLS) with each clip's calibration and refining its outline against the images.

\paragraph{Camera Matching and Harmonization.}
Before harmonization, we match the rendered object to the camera by applying a Gaussian blur with $\sigma = 0.55$\,px, adding Gaussian noise whose standard deviation is estimated from the surrounding image region (capped at 1.5 intensity levels), and applying bounded corrections of white balance, highlight level and saturation towards a ring of pixels around the object. PCT-Net~\cite{10204242}, as implemented in the libcom library~\cite{niu2021making}, is then applied only to the object, excluding its shadow, and all pixels outside the object are restored from its input. On \gls{reactpai}, PCT-Net operates on a crop around the object, padded by the object's size; objects with fewer than 256 pixels or a bounding box smaller than $12 \times 24$ pixels are not harmonized, the output is blended 50/50 with the input, and it is discarded if PCT-Net changes the object's pixels by more than 24 intensity levels on average or the blend has less than half of the object's original luminance contrast. Finally, we automatically verify that every edited image differs from the original only within the mask of the inserted object and its shadow.

\begin{figure*}[t]
    \centering
    \includegraphics[width=\linewidth]{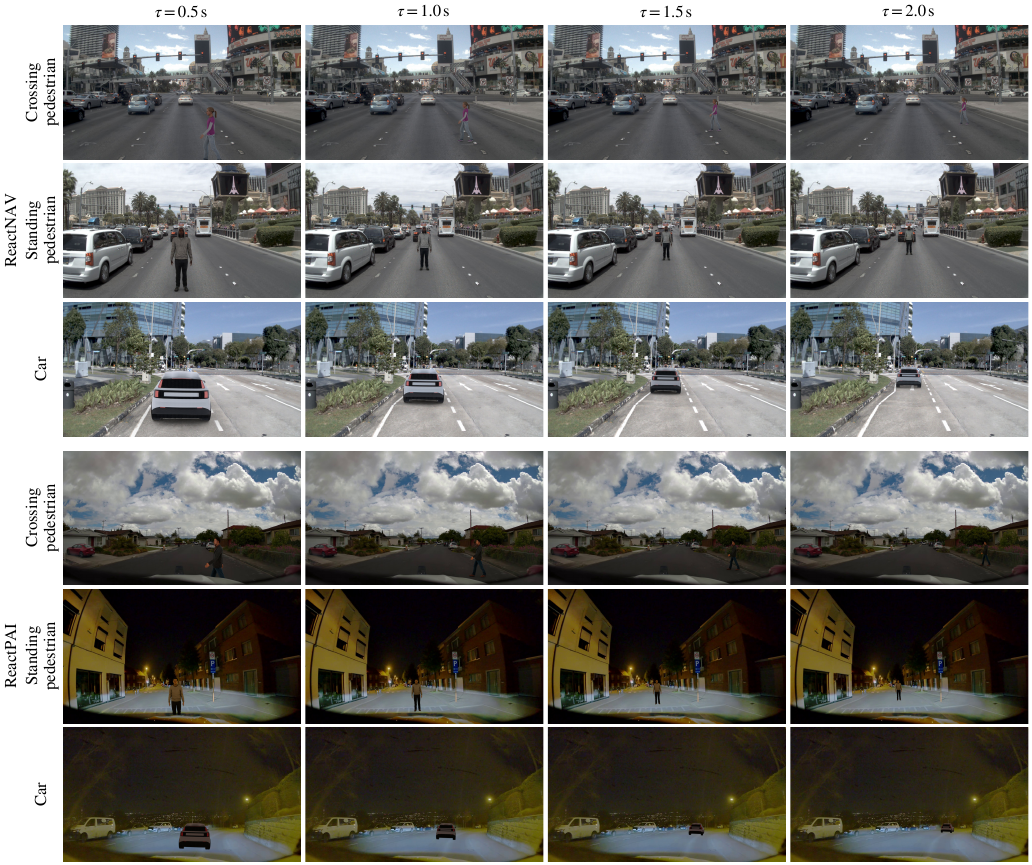}

    \caption{Randomly selected \gls{reactnav} (top) and \gls{reactpai} (bottom) samples for each inserted object and look-ahead time $\tau$, shown in the edited front-camera image at the current frame $t$. Each row shows a single scene, whose four look-ahead variants share the same asset.}
    \label{fig:reactbench_examples}
\end{figure*}

\paragraph{Examples.} \Cref{fig:reactbench_examples} depicts randomly selected samples from both \gls{reactnav} and \gls{reactpai} for each inserted object and look-ahead time $\tau$.

\paragraph{Vehicle Asset Attribution.} The seven vehicle assets come from OpenX Assets~\cite{vevalabs2025openxassets}, which adapts Sketchfab models licensed under the Creative Commons Attribution 4.0 International licence (CC~BY~4.0): Audi TT Roadster (2014), Dacia Duster (2010), Hyundai Tucson (2015) and Mini Countryman (2016) by razor24; BMW X1 (2016) by mk2design; Volvo EX30 (2024) by Mehdi Lagzouli (LagzDesign); and Volvo V60 Polestar (2013) by myedsu.

\paragraph{Availability.}
We will make the compositing code and the edited \gls{reactnav} samples publicly available. The \gls{paiav} license does not permit redistributing modified data, so the edited \gls{reactpai} samples may not be released directly; in that case, we will instead provide the identifiers of the selected clips together with the scripts that regenerate the edited samples from the original \gls{paiav} data.

\subsection{Could Compositing Explain the Failures?}
\label{subsec:supp_detection_check}
Since the inserted objects are not fully photorealistic, the models might fail to react to them because they do not recognize them or dismiss them as artifacts, rather than because they cannot handle the hazard. We address this concern in two ways: an off-the-shelf detector recognizes almost all inserted objects, and the single-frame models also fail to brake for real stationary vehicles.

\paragraph{Detectability.}
We run D-FINE-X~\cite{ICLR2025_6cf58a87}, pretrained on Objects365 and fine-tuned on COCO, on the edited front-camera image at the current frame $t$, at its default $640\times640$ input resolution. Since each edited image differs from its original only in the object and its shadow, we take as the target box the tight box around pixels that change by more than 40 gray levels inside the object's projected 3D bounding box. The threshold removes the faint shadow, and the projected box removes the shadow outside it. A sample counts as detected if a \texttt{person} detection (for pedestrians) or a \texttt{car} detection (for cars) with score $\geq 0.5$ overlaps the target box with IoU $\geq 0.5$ in the edited image, and no such detection exists in the original image. We exclude the 158 of 51{,}276 samples in which the object is not visible in the front camera at $t$. As \cref{tab:detection_check} shows, the detector finds the inserted object in 99.3\% of the remaining samples. Detection rates drop slightly with $\tau$, i.e., for smaller and more distant objects, and are lower for \gls{reactpai}, whose wide-angle front camera renders objects smaller and which contains more night scenes. Most remaining misses are objects that are only partially in view or poorly lit.

\paragraph{Real Stationary Vehicles.}
We further evaluate the models on the real-scenes test set of \cref{subsec:sft_on_artificial_targets}, in which the hazard is a real stationary vehicle and only the ego status is edited (Appendix~\ref{subsec:supp_generalization_sets}). We restrict this control to ReCogDrive, DriveWAM, and SimWAM, which receive only the current frame, so that the edited ego status does not contradict their visual input. As \cref{tab:real_scene_control} shows, none of them produces a safe plan at $\tau \leq 1.0\,\mathrm{s}$, and at most 1.0\% of their plans are safe at $\tau = 1.5\,\mathrm{s}$, where a deceleration of at most $3.0\,\mathrm{m/s^2}$ would suffice to stop before the vehicle. Only at $\tau = 2.0\,\mathrm{s}$, where the edited speeds are low (median 12\,km/h), are 23--29\% of the plans safe.

\begin{table}[t]
    \centering
    \small
    \setlength{\tabcolsep}{5pt}
    \caption{Share of edited samples (\%) in which D-FINE-X detects the inserted object at its location, by object and look-ahead time $\tau$.}
    \label{tab:detection_check}
    \begin{tabular}{lccccc}
        \toprule
        & \multicolumn{4}{c}{$\tau$ (s)} & \\
        \cmidrule(lr){2-5}
        Object & 0.5 & 1.0 & 1.5 & 2.0 & All \\
        \midrule
        \multicolumn{6}{l}{\textit{\gls{reactnav}}} \\
        Crossing ped. & 99.9  & 99.9  & 99.5 & 99.3 & 99.7 \\
        Standing ped. & 100.0 & 100.0 & 99.9 & 99.8 & 99.9 \\
        Car           & 100.0 & 100.0 & 99.8 & 99.6 & 99.8 \\
        \midrule
        \multicolumn{6}{l}{\textit{\gls{reactpai}}} \\
        Crossing ped. & 98.5 & 98.3 & 97.3 & 95.8 & 97.5 \\
        Standing ped. & 99.0 & 98.6 & 97.1 & 95.7 & 97.6 \\
        Car           & 99.0 & 98.8 & 98.5 & 98.1 & 98.6 \\
        \bottomrule
    \end{tabular}
\end{table}

\begin{table}[t]
    \centering
    \caption{$P$ (in percent) on the 308 real scenes with a stationary vehicle ahead, by look-ahead time $\tau$ (s).}
    \label{tab:real_scene_control}
    \footnotesize
    \begin{tabular}{@{}l*{5}{r}@{}}
        \toprule
        Model & $\tau{=}0.5$ & 1.0 & 1.5 & 2.0 & Mean \\
        \midrule
        ReCogDrive & 0.0 & 0.0 & 1.0 & 22.7 & 5.9 \\
        DriveWAM & 0.0 & 0.0 & 0.6 & 28.9 & 7.4 \\
        SimWAM & 0.0 & 0.0 & 0.0 & 23.4 & 5.8 \\
        \bottomrule
    \end{tabular}
\end{table}

\subsection{Data Statistics}
\label{subsec:supp_data_stats}
\Cref{tab:speed_accel_distribution} shows the number of \gls{reactnav} and \gls{reactpai} samples in each ego-vehicle speed interval at the current frame; \cref{fig:deltaP_by_vel} (Appendix~\ref{subsec:supp_results}) shows how the safety improvement varies with this speed.

\begin{table}[t]
    \centering
    \caption{Number of samples in each ego-vehicle speed interval at the current frame for \gls{reactnav} and \gls{reactpai}.}
    \label{tab:speed_accel_distribution}
    \footnotesize
    \setlength{\tabcolsep}{3pt}
    \begin{tabular}{@{}l*{8}{r}@{}}
        \toprule
        km/h & 5--10 & 10--15 & 15--20 & 20--25 & 25--30 & 30--35 & 35--40 & Total \\
        \midrule
        ReactNAV & 325 & 634 & 803 & 680 & 484 & 271 & 31  & 3,228 \\
        ReactPAI & 180 & 188 & 199 & 137 & 127 & 110 & 104 & 1,045 \\
        \bottomrule
    \end{tabular}
\end{table}

\subsection{Metrics and Trajectory Interpolation}
\label{subsec:supp_metrics_trajectory}
This section complements \cref{subsec:benchmark_details} with implementation details of the evaluation metrics and the trajectory interpolation.

\paragraph{Footprints.}
Collisions are checked between rectangular footprints in the ground plane. The ego footprint is placed relative to the rear axle and oriented by the predicted heading; its dimensions are NAVSIM's ego vehicle parameters on \gls{reactnav} and the vehicle dimensions provided with each clip on \gls{reactpai}. Recorded road users use their annotated bounding boxes. For each pedestrian asset, we measure a single bounding box as the union of the bounds of its meshes over its entire walk cycle, so that it encloses every pose of the asset. Its 3D extent is used for the placement validity check of Appendix~\ref{subsec:supp_object_compositing}, and its ground-plane projection for collision checking. The inserted car uses the bounding box provided with the asset.

\paragraph{Scene.}
As in NAVSIM's non-reactive evaluation, recorded road users follow their logged motion and do not react to the ego vehicle. On \gls{reactnav}, they are annotated every 0.5\,s; on \gls{reactpai}, we use the automatically generated obstacle labels provided with the dataset, which are available at about 10\,Hz and which we transform into the ego frame at the current frame $t$ using the logged ego motion. The inserted object moves at constant velocity over the whole horizon: the crossing pedestrian keeps walking at 1.4\,m/s after crossing the ego vehicle's path, and the standing pedestrian and the car remain stationary.

\paragraph{At-fault classification.}
We follow the at-fault criteria of NAVSIM with three modifications: (a) a lateral contact with a pedestrian or cyclist while the ego vehicle is moving is considered at fault, (b) a contact that occurs while the ego vehicle is moving backwards (i.e., its displacement is opposite to its predicted heading) is never exempted as a rear-end collision, and (c) collisions with static objects also score 0 instead of 0.5. Only the first contact with each road user is considered, as in PDMS. The ego speed at a contact is the speed of the predicted trajectory segment leading into it, and a road user's speed is its displacement over a 1\,s window around the contact; inserted objects use their known velocity, and static objects (e.g., barriers and traffic cones) are always treated as stopped. Each contact is assigned to the first of the following classes that applies:
\begin{enumerate}
    \item the ego vehicle is stopped ($\leq 0.05$\,m/s): not at fault;
    \item the other road user is stopped ($\leq 0.05$\,m/s): at fault;
    \item rear contact, \ie, the road user's center lies more than $150^\circ$ from the ego heading as seen from the rear axle: not at fault, unless the ego vehicle is moving backwards (modification (b));
    \item frontal contact, \ie, the front edge of the ego footprint touches the road user: at fault;
    \item lateral contact: at fault if the road user is a pedestrian or cyclist (modification (a)), if the ego vehicle occupies more than one lane, \ie, more than one lane polygon contains an ego corner and none contains all four, or if any ego corner lies outside the drivable area.
\end{enumerate}
Since \gls{paiav} provides no map, lateral contacts with vehicles are never at fault on \gls{reactpai}. This choice is immaterial: counting all of them as at fault instead changes the outcome of at most one of the 12,540 edited predictions per model.

\paragraph{Inserted-object collisions.}
$\NCins$ is computed like $\NC$ but considers only at-fault collisions with the inserted object. Since such a collision also sets $P$ to zero, $\NCins - P$ is the fraction of trajectories that avoid the inserted object but collide with another road user or, on \gls{reactnav}, leave the drivable area. Hence, $\dNCins$ measures the edit effect on collisions with the inserted object alone, whereas $\dP$ also reflects changes in collisions with other road users and in drivable area compliance.

\paragraph{Contact speed and ADE.}
The contact speed used in $\CSC$ is the speed of the ego rear axle along the predicted segment leading into the first at-fault contact. It measures how fast the ego vehicle is driving, not the relative impact speed, and the two trajectories of a pair may first collide with different road users. The \gls{ade} used in $\TIR$ is the mean Euclidean distance between corresponding waypoints of $\mathbf{y}^{\mathrm{O}}$ and $\mathbf{y}^{\mathrm{E}}$ over the first 4\,s, at each model's native resolution (40 waypoints for the Alpamayo models and 8 for the others); a pair is counted as invariant if this distance is strictly below $\epsilon$.

\paragraph{Trajectory interpolation.}
Driving models typically predict waypoints at a low temporal resolution (\eg, 2\,Hz), and checking only at the waypoints can miss collisions that occur between them or misclassify the type of contact, and hence fault. Rather than tracking the waypoints with NAVSIM's LQR controller, which not all evaluated models were trained with, we interpolate the predicted positions and (unwrapped) headings piecewise-linearly, starting from the ego vehicle's current pose, \ie, with constant velocity between consecutive waypoints. This interpolant passes exactly through the waypoints, has no parameters, and never overshoots. We interpolate the annotated agent tracks in the same way within their annotated lifetimes. We check collisions at 100\,Hz and refine each first contact to the exact instant of first touch, at which the contact is classified. Drivable area compliance is checked at 10\,Hz, as in NAVSIM, whereas $\CSC$ and $\TIR$ are computed at each model's native waypoint frequency.

\section{Further Experimental Details and Results}

\subsection{Evaluated Models}
\label{subsec:supp_models}
This section complements \cref{subsec:models} with details of the evaluated models and of how we adapt their inputs to \benchname. \Cref{tab:model_comparison} summarizes their action heads, input configurations, and outputs.

\begin{table*}[t]
\centering
\caption{Comparison of the action heads, input configurations, and outputs of the evaluated models. Visual and ego history denote the temporal spans of camera observations and past ego states, respectively; a visual history of 0\,s indicates a single current frame.}
\label{tab:model_comparison}
\small
\renewcommand{\arraystretch}{1.15}
\setlength{\tabcolsep}{3pt}
    \begin{tabular*}{\textwidth}{@{\extracolsep{\fill}}llccccccc@{}}
    \toprule
    \multirow{2}{*}{Model}
    & \multirow{2}{*}{Action head}
    & \multicolumn{5}{c}{Inputs}
    & \multicolumn{2}{c}{Outputs} \\
    \cmidrule(lr){3-7}
    \cmidrule(lr){8-9}
    & & Cameras
    & \shortstack{Frames\\per camera}
    & \shortstack{Visual history\\(s)}
    & \shortstack{Ego history\\(s)}
    & Vel./acc.
    & Reasoning
    & \shortstack{Trajectory\\horizon (s)} \\
    \midrule
    Alpamayo~1.5
    & Transformer Decoder
    & 4 & 4 & 0.3 & 1.5 & No & Yes & 6.4 \\

    Alpamayo~2
    & Transformer Decoder
    & 6 & 4 & 0.3 & 1.5 & No & Yes & 6.4 \\

    AutoVLA
    & --
    & 3 & 4 & 1.5 & -- & Yes & Optional & 5.0 \\

    ReCogDrive
    & Diffusion Transformer
    & 1 & 1 & 0 & 1.5 & Yes & No & 4.0 \\

    DriveWAM
    & --
    & 1 & 1 & 0 & 1.5 & Yes & No & 4.0 \\

    SimWAM
    & Diffusion Transformer
    & 1 & 1 & 0 & -- & Yes & No & 4.0 \\
    \bottomrule
    \end{tabular*}
\end{table*}

\paragraph{Action Heads.}
AutoVLA and DriveWAM do not introduce a dedicated action module: AutoVLA adds 2,048 action tokens to the existing language vocabulary and generates the trajectory autoregressively as a sequence of ten such tokens, whereas DriveWAM places action tokens in the same sequence as the video tokens and denoises them with flow matching, using the same diffusion transformer that generates the future video frames.

\paragraph{Training Data.}
AutoVLA, ReCogDrive, DriveWAM, and SimWAM were all trained on NAVSIM data, and hence their expected inputs follow the NAVSIM format. The Alpamayo models, in contrast, were trained primarily on NVIDIA's multi-camera driving data (80k hours for Alpamayo~1.5 and roughly 115k hours for Alpamayo~2), with \gls{paiav} listed among their training datasets, and their input format follows \gls{paiav}. Alpamayo~1.5 expects frames 0.1\,s apart from four cameras: front-wide, front-tele, cross-left, and cross-right. Alpamayo~2 uses the same four cameras plus two rear-facing ones, rear-left and rear-right. Per its model card~\cite{nvidia2026alpamayo15}, Alpamayo~1.5 was additionally trained on a range of public driving datasets, including NAVSIM-derived data; no comparable information is publicly available for Alpamayo~2~\cite{nvidia2026alpamayo2}. Thus, although NAVSIM is not entirely out-of-distribution for Alpamayo~1.5, \gls{paiav} is considerably closer to its training distribution.

\paragraph{Input Adaptation and Evaluation.}
Because the models predict trajectories of different lengths, ranging from 4.0 to 6.4\,s into the future (\cref{tab:model_comparison}), we compute the metrics defined in \cref{subsec:benchmark_details} using only the first 4\,s of each predicted trajectory for a fair comparison; that is, we discard the last 1.0 and 2.4\,s of the trajectories predicted by AutoVLA and the two Alpamayo models, respectively. For every model, the original and edited predictions of a sample use the same random seed, derived from the sample identifier, for text generation and trajectory sampling where applicable. Differences between $\mathbf{y}^{\mathrm{O}}$ and $\mathbf{y}^{\mathrm{E}}$ are thus caused by the edit rather than by independent resampling. When evaluating the Alpamayo models on \gls{reactnav}, we use only the three front-facing NAVSIM cameras, which we map to their \gls{paiav} counterparts, and drop the remaining \gls{paiav} camera inputs, namely the front-tele camera and, for Alpamayo~2, also the two rear cameras. Moreover, because NAVSIM frames are 0.5\,s apart, the visual history spans 1.5\,s instead of 0.3\,s, and we linearly interpolate the four NAVSIM ego poses to the 16 poses at 10\,Hz that the models expect. Finally, the two subsets differ in the navigation input: on \gls{reactnav}, we pass NAVSIM's driving command to both models as a navigation instruction (\eg, ``Turn left at the next opportunity.''), whereas on \gls{reactpai}, for which no navigation commands are available, they receive none. On \gls{reactnav}, AutoVLA occasionally generated a number of action tokens other than the expected ten and thus no valid trajectory; we exclude the affected 1--5 samples per inserted object and look-ahead time (at most 0.15\%). Moreover, for NAVSIM, AutoVLA's front-left and front-right inputs are the side cameras \texttt{CAM\_L1} and \texttt{CAM\_R1}, which we do not edit; Appendix~\ref{subsec:supp_autovla_cameras} repeats the evaluation with the cameras described in the AutoVLA paper. We keep them unedited and, for AutoVLA only, additionally exclude the 1, 4, and 9 crossing-pedestrian samples at $\tau = 1.0$, $1.5$, and $2.0\,\mathrm{s}$ in which the pedestrian would enter the field of view of these side cameras. All other models are evaluated on all 3,228 samples.

\subsection{Results}
\label{subsec:supp_results}

\Cref{tab:main_results_detailed} is the more detailed version of \cref{tab:main_results}, with the original and edited version of each metric along with their differences, and additionally reports $\NCins$ (Appendix~\ref{subsec:supp_metrics_trajectory}).

\begin{table*}[p]
    \centering
    \caption{Detailed results on ReactNAV and ReactPAI by model, inserted object, and look-ahead time $\tau$ (s). O and E denote the trajectories predicted from the original and edited scene, respectively, both evaluated in the edited scene, and $\Delta = \mathrm{E} - \mathrm{O}$. $P$ and $\NCins$ are in percent and their deltas in percentage points (higher is better). $\CS$ is the mean contact speed in km/h over the samples for which both trajectories result in an at-fault collision (lower is better). Deltas are computed before rounding.}
    \label{tab:main_results_detailed}
    \scriptsize
    \setlength{\tabcolsep}{3.4pt}
    \renewcommand{\arraystretch}{0.88}
    \begin{tabular}{llc*{4}{r}@{\hspace{12pt}}*{4}{r}@{\hspace{12pt}}*{4}{r}}
        \toprule
        & & & \multicolumn{4}{c}{Crossing pedestrian} & \multicolumn{4}{c}{Standing pedestrian} & \multicolumn{4}{c}{Car} \\
        \cmidrule(lr){4-7} \cmidrule(lr){8-11} \cmidrule(lr){12-15}
        Model & Metric & & $\tau{=}0.5$ & 1.0 & 1.5 & 2.0 & 0.5 & 1.0 & 1.5 & 2.0 & 0.5 & 1.0 & 1.5 & 2.0 \\
        \midrule
        \multicolumn{15}{l}{\textit{ReactNAV}} \\
        \midrule
        Alpamayo~1.5 & \multirow{3}{*}{$\boldsymbol{P}$} & O & 0.00 & 0.06 & 2.11 & 5.89 & 0.00 & 0.03 & 1.55 & 5.11 & 0.00 & 0.00 & 1.15 & 3.69 \\
         &  & E & 0.25 & 3.16 & 9.26 & 14.59 & 0.09 & 3.69 & 13.48 & 22.74 & 0.12 & 3.84 & 17.07 & 32.50 \\
        \rowcolor{gray!15} \cellcolor{white} & \cellcolor{white} & $\Delta$ & $+0.25$ & $+3.10$ & $+7.16$ & $+8.71$ & $+0.09$ & $+3.66$ & $+11.93$ & $+17.63$ & $+0.12$ & $+3.84$ & $+15.92$ & $+28.81$ \\
        \addlinespace[0.5pt]
         & \multirow{3}{*}{$\NCins$} & O & 0.00 & 0.15 & 2.94 & 9.23 & 0.00 & 0.19 & 3.22 & 11.06 & 0.00 & 0.03 & 1.86 & 6.32 \\
         &  & E & 0.25 & 3.44 & 10.90 & 18.68 & 0.09 & 4.28 & 17.72 & 31.44 & 0.12 & 4.06 & 18.62 & 37.08 \\
        \rowcolor{gray!15} \cellcolor{white} & \cellcolor{white} & $\Delta$ & $+0.25$ & $+3.28$ & $+7.96$ & $+9.45$ & $+0.09$ & $+4.09$ & $+14.50$ & $+20.38$ & $+0.12$ & $+4.03$ & $+16.76$ & $+30.76$ \\
        \addlinespace[0.5pt]
         & \multirow{3}{*}{$\CS$} & O & 18.93 & 18.86 & 18.79 & 18.61 & 18.93 & 18.95 & 19.06 & 18.87 & 18.91 & 18.94 & 19.38 & 20.13 \\
         &  & E & 17.42 & 16.14 & 16.11 & 16.32 & 17.30 & 14.43 & 13.62 & 14.04 & 17.68 & 14.65 & 12.33 & 12.88 \\
        \rowcolor{gray!15} \cellcolor{white} & \cellcolor{white} & $\Delta$ & $-1.51$ & $-2.73$ & $-2.68$ & $-2.29$ & $-1.63$ & $-4.52$ & $-5.44$ & $-4.83$ & $-1.23$ & $-4.29$ & $-7.05$ & $-7.25$ \\
        \addlinespace[2pt]
        Alpamayo~2 & \multirow{3}{*}{$\boldsymbol{P}$} & O & 0.00 & 0.34 & 3.75 & 10.53 & 0.00 & 0.46 & 4.28 & 11.18 & 0.00 & 0.34 & 3.59 & 10.19 \\
         &  & E & 0.06 & 2.45 & 10.50 & 18.90 & 0.03 & 3.72 & 13.69 & 23.51 & 0.12 & 5.70 & 22.99 & 32.62 \\
        \rowcolor{gray!15} \cellcolor{white} & \cellcolor{white} & $\Delta$ & $+0.06$ & $+2.11$ & $+6.75$ & $+8.36$ & $+0.03$ & $+3.25$ & $+9.42$ & $+12.33$ & $+0.12$ & $+5.36$ & $+19.39$ & $+22.43$ \\
        \addlinespace[0.5pt]
         & \multirow{3}{*}{$\NCins$} & O & 0.00 & 2.70 & 17.87 & 35.56 & 0.00 & 4.71 & 25.50 & 43.43 & 0.00 & 1.89 & 16.64 & 35.81 \\
         &  & E & 0.06 & 5.36 & 21.44 & 38.79 & 0.03 & 10.35 & 34.17 & 52.04 & 0.15 & 9.08 & 39.28 & 59.05 \\
        \rowcolor{gray!15} \cellcolor{white} & \cellcolor{white} & $\Delta$ & $+0.06$ & $+2.66$ & $+3.56$ & $+3.22$ & $+0.03$ & $+5.64$ & $+8.67$ & $+8.61$ & $+0.15$ & $+7.19$ & $+22.65$ & $+23.23$ \\
        \addlinespace[0.5pt]
         & \multirow{3}{*}{$\CS$} & O & 18.82 & 18.66 & 19.00 & 19.53 & 18.85 & 18.62 & 19.02 & 19.60 & 18.80 & 18.71 & 19.69 & 20.72 \\
         &  & E & 17.71 & 16.89 & 17.58 & 18.79 & 17.36 & 15.53 & 15.84 & 17.11 & 17.61 & 15.76 & 15.52 & 16.50 \\
        \rowcolor{gray!15} \cellcolor{white} & \cellcolor{white} & $\Delta$ & $-1.11$ & $-1.77$ & $-1.42$ & $-0.73$ & $-1.49$ & $-3.10$ & $-3.18$ & $-2.50$ & $-1.19$ & $-2.96$ & $-4.17$ & $-4.22$ \\
        \addlinespace[2pt]
        AutoVLA & \multirow{3}{*}{$\boldsymbol{P}$} & O & 0.00 & 0.03 & 0.09 & 0.31 & 0.00 & 0.03 & 0.09 & 0.31 & 0.00 & 0.09 & 0.09 & 0.50 \\
        &  & E & 0.00 & 0.00 & 0.62 & 1.21 & 0.00 & 0.19 & 0.43 & 1.27 & 0.03 & 0.96 & 2.23 & 3.75 \\
        \rowcolor{gray!15} \cellcolor{white} & \cellcolor{white} & $\Delta$ & $+0.00$ & $-0.03$ & $+0.53$ & $+0.90$ & $+0.00$ & $+0.16$ & $+0.34$ & $+0.96$ & $+0.03$ & $+0.87$ & $+2.14$ & $+3.26$ \\
        \addlinespace[0.5pt]
        & \multirow{3}{*}{$\NCins$} & O & 0.00 & 0.06 & 0.22 & 0.53 & 0.00 & 0.12 & 0.25 & 0.56 & 0.00 & 0.12 & 0.25 & 0.68 \\
        &  & E & 0.00 & 0.09 & 0.96 & 1.71 & 0.03 & 0.34 & 0.93 & 1.83 & 0.03 & 1.15 & 2.70 & 4.56 \\
        \rowcolor{gray!15} \cellcolor{white} & \cellcolor{white} & $\Delta$ & $+0.00$ & $+0.03$ & $+0.74$ & $+1.18$ & $+0.03$ & $+0.22$ & $+0.68$ & $+1.27$ & $+0.03$ & $+1.02$ & $+2.45$ & $+3.88$ \\
        \addlinespace[0.5pt]
        & \multirow{3}{*}{$\CS$} & O & 18.86 & 18.64 & 18.44 & 18.24 & 18.86 & 18.63 & 18.44 & 18.23 & 18.83 & 18.68 & 18.55 & 18.47 \\
        &  & E & 18.89 & 18.62 & 18.42 & 18.21 & 18.85 & 18.55 & 18.27 & 18.07 & 18.55 & 18.09 & 17.85 & 17.80 \\
        \rowcolor{gray!15} \cellcolor{white} & \cellcolor{white} & $\Delta$ & $+0.03$ & $-0.02$ & $-0.02$ & $-0.03$ & $-0.01$ & $-0.08$ & $-0.16$ & $-0.16$ & $-0.28$ & $-0.59$ & $-0.70$ & $-0.67$ \\
        \addlinespace[2pt]
        ReCogDrive & \multirow{3}{*}{$\boldsymbol{P}$} & O & 0.00 & 0.03 & 0.00 & 0.34 & 0.00 & 0.00 & 0.00 & 0.34 & 0.00 & 0.00 & 0.00 & 0.06 \\
         &  & E & 0.00 & 0.00 & 0.22 & 0.74 & 0.00 & 0.00 & 0.12 & 0.74 & 0.00 & 0.00 & 0.28 & 1.95 \\
        \rowcolor{gray!15} \cellcolor{white} & \cellcolor{white} & $\Delta$ & $+0.00$ & $-0.03$ & $+0.22$ & $+0.40$ & $+0.00$ & $+0.00$ & $+0.12$ & $+0.40$ & $+0.00$ & $+0.00$ & $+0.28$ & $+1.89$ \\
        \addlinespace[0.5pt]
         & \multirow{3}{*}{$\NCins$} & O & 0.00 & 0.03 & 0.06 & 0.50 & 0.00 & 0.00 & 0.03 & 0.50 & 0.00 & 0.00 & 0.03 & 0.12 \\
         &  & E & 0.00 & 0.00 & 0.28 & 0.90 & 0.00 & 0.00 & 0.22 & 1.05 & 0.00 & 0.00 & 0.43 & 2.23 \\
        \rowcolor{gray!15} \cellcolor{white} & \cellcolor{white} & $\Delta$ & $+0.00$ & $-0.03$ & $+0.22$ & $+0.40$ & $+0.00$ & $+0.00$ & $+0.19$ & $+0.56$ & $+0.00$ & $+0.00$ & $+0.40$ & $+2.11$ \\
        \addlinespace[0.5pt]
         & \multirow{3}{*}{$\CS$} & O & 17.82 & 19.98 & 20.19 & 19.78 & 17.61 & 19.73 & 20.08 & 19.84 & 17.97 & 20.09 & 20.17 & 19.90 \\
         &  & E & 18.13 & 19.52 & 19.61 & 19.33 & 18.13 & 19.46 & 19.02 & 18.63 & 19.39 & 20.02 & 19.21 & 18.82 \\
        \rowcolor{gray!15} \cellcolor{white} & \cellcolor{white} & $\Delta$ & $+0.31$ & $-0.45$ & $-0.57$ & $-0.45$ & $+0.51$ & $-0.26$ & $-1.06$ & $-1.20$ & $+1.42$ & $-0.07$ & $-0.96$ & $-1.08$ \\
        \addlinespace[2pt]
        DriveWAM & \multirow{3}{*}{$\boldsymbol{P}$} & O & 0.00 & 0.00 & 0.03 & 0.22 & 0.00 & 0.00 & 0.00 & 0.15 & 0.00 & 0.00 & 0.03 & 0.15 \\
         &  & E & 0.00 & 0.06 & 0.15 & 0.31 & 0.00 & 0.12 & 0.56 & 2.23 & 0.00 & 1.77 & 3.66 & 5.76 \\
        \rowcolor{gray!15} \cellcolor{white} & \cellcolor{white} & $\Delta$ & $+0.00$ & $+0.06$ & $+0.12$ & $+0.09$ & $+0.00$ & $+0.12$ & $+0.56$ & $+2.08$ & $+0.00$ & $+1.77$ & $+3.62$ & $+5.61$ \\
        \addlinespace[0.5pt]
         & \multirow{3}{*}{$\NCins$} & O & 0.00 & 0.03 & 0.06 & 0.28 & 0.00 & 0.03 & 0.03 & 0.22 & 0.00 & 0.03 & 0.06 & 0.19 \\
         &  & E & 0.00 & 0.09 & 0.19 & 0.34 & 0.00 & 0.15 & 0.59 & 2.35 & 0.00 & 1.77 & 3.69 & 5.82 \\
        \rowcolor{gray!15} \cellcolor{white} & \cellcolor{white} & $\Delta$ & $+0.00$ & $+0.06$ & $+0.12$ & $+0.06$ & $+0.00$ & $+0.12$ & $+0.56$ & $+2.14$ & $+0.00$ & $+1.73$ & $+3.62$ & $+5.64$ \\
        \addlinespace[0.5pt]
         & \multirow{3}{*}{$\CS$} & O & 18.89 & 18.70 & 18.53 & 18.33 & 18.89 & 18.71 & 18.59 & 18.57 & 18.78 & 18.82 & 18.86 & 18.92 \\
         &  & E & 18.81 & 18.56 & 18.36 & 18.18 & 18.74 & 18.29 & 17.87 & 17.69 & 18.19 & 17.94 & 17.96 & 18.16 \\
        \rowcolor{gray!15} \cellcolor{white} & \cellcolor{white} & $\Delta$ & $-0.08$ & $-0.14$ & $-0.17$ & $-0.14$ & $-0.15$ & $-0.43$ & $-0.72$ & $-0.88$ & $-0.60$ & $-0.88$ & $-0.91$ & $-0.76$ \\
        \addlinespace[2pt]
        SimWAM & \multirow{3}{*}{$\boldsymbol{P}$} & O & 0.00 & 0.00 & 0.03 & 0.34 & 0.00 & 0.00 & 0.03 & 0.40 & 0.00 & 0.00 & 0.03 & 0.12 \\
         &  & E & 0.00 & 0.06 & 0.15 & 0.56 & 0.00 & 0.00 & 0.19 & 1.12 & 0.00 & 0.06 & 1.02 & 2.39 \\
        \rowcolor{gray!15} \cellcolor{white} & \cellcolor{white} & $\Delta$ & $+0.00$ & $+0.06$ & $+0.12$ & $+0.22$ & $+0.00$ & $+0.00$ & $+0.15$ & $+0.71$ & $+0.00$ & $+0.06$ & $+0.99$ & $+2.26$ \\
        \addlinespace[0.5pt]
         & \multirow{3}{*}{$\NCins$} & O & 0.00 & 0.00 & 0.06 & 0.40 & 0.00 & 0.03 & 0.06 & 0.43 & 0.00 & 0.00 & 0.06 & 0.15 \\
         &  & E & 0.00 & 0.06 & 0.19 & 0.62 & 0.00 & 0.03 & 0.25 & 1.24 & 0.00 & 0.06 & 1.05 & 2.57 \\
        \rowcolor{gray!15} \cellcolor{white} & \cellcolor{white} & $\Delta$ & $+0.00$ & $+0.06$ & $+0.12$ & $+0.22$ & $+0.00$ & $+0.00$ & $+0.19$ & $+0.81$ & $+0.00$ & $+0.06$ & $+0.99$ & $+2.42$ \\
        \addlinespace[0.5pt]
         & \multirow{3}{*}{$\CS$} & O & 19.99 & 20.22 & 19.78 & 18.79 & 19.96 & 19.64 & 19.74 & 18.82 & 19.90 & 21.14 & 20.26 & 19.17 \\
         &  & E & 19.88 & 20.07 & 18.96 & 18.47 & 19.89 & 19.43 & 18.10 & 17.56 & 20.35 & 20.34 & 18.47 & 18.37 \\
        \rowcolor{gray!15} \cellcolor{white} & \cellcolor{white} & $\Delta$ & $-0.11$ & $-0.14$ & $-0.82$ & $-0.32$ & $-0.07$ & $-0.21$ & $-1.64$ & $-1.26$ & $+0.45$ & $-0.79$ & $-1.79$ & $-0.79$ \\
        \midrule
        \multicolumn{15}{l}{\textit{ReactPAI}} \\
        \midrule
        Alpamayo~1.5 & \multirow{3}{*}{$\boldsymbol{P}$} & O & 0.00 & 0.00 & 0.67 & 2.68 & 0.00 & 0.00 & 0.19 & 2.30 & 0.00 & 0.00 & 0.67 & 1.72 \\
         &  & E & 0.10 & 5.55 & 25.84 & 37.80 & 0.19 & 5.07 & 21.82 & 37.99 & 0.67 & 12.92 & 21.53 & 27.37 \\
        \rowcolor{gray!15} \cellcolor{white} & \cellcolor{white} & $\Delta$ & $+0.10$ & $+5.55$ & $+25.17$ & $+35.12$ & $+0.19$ & $+5.07$ & $+21.63$ & $+35.69$ & $+0.67$ & $+12.92$ & $+20.86$ & $+25.65$ \\
        \addlinespace[0.5pt]
         & \multirow{3}{*}{$\NCins$} & O & 0.00 & 0.00 & 0.67 & 2.78 & 0.00 & 0.00 & 0.19 & 2.39 & 0.00 & 0.00 & 0.67 & 1.72 \\
         &  & E & 0.10 & 5.55 & 25.84 & 37.99 & 0.19 & 5.26 & 21.82 & 38.09 & 0.67 & 13.01 & 21.72 & 27.66 \\
        \rowcolor{gray!15} \cellcolor{white} & \cellcolor{white} & $\Delta$ & $+0.10$ & $+5.55$ & $+25.17$ & $+35.22$ & $+0.19$ & $+5.26$ & $+21.63$ & $+35.69$ & $+0.67$ & $+13.01$ & $+21.05$ & $+25.93$ \\
        \addlinespace[0.5pt]
         & \multirow{3}{*}{$\CS$} & O & 20.33 & 20.57 & 21.01 & 21.63 & 20.35 & 20.53 & 20.79 & 21.31 & 20.35 & 21.52 & 22.30 & 22.59 \\
         &  & E & 16.69 & 12.61 & 13.50 & 15.87 & 16.22 & 11.39 & 10.53 & 12.52 & 15.92 & 12.44 & 12.57 & 14.07 \\
        \rowcolor{gray!15} \cellcolor{white} & \cellcolor{white} & $\Delta$ & $-3.64$ & $-7.96$ & $-7.52$ & $-5.75$ & $-4.13$ & $-9.14$ & $-10.25$ & $-8.79$ & $-4.43$ & $-9.08$ & $-9.73$ & $-8.52$ \\
        \addlinespace[2pt]
        Alpamayo~2 & \multirow{3}{*}{$\boldsymbol{P}$} & O & 0.00 & 0.10 & 0.29 & 2.11 & 0.00 & 0.10 & 0.19 & 2.20 & 0.00 & 0.10 & 0.57 & 2.01 \\
         &  & E & 0.00 & 3.54 & 22.97 & 36.17 & 0.00 & 0.67 & 8.80 & 18.09 & 0.10 & 5.65 & 17.22 & 27.18 \\
        \rowcolor{gray!15} \cellcolor{white} & \cellcolor{white} & $\Delta$ & $+0.00$ & $+3.44$ & $+22.68$ & $+34.07$ & $+0.00$ & $+0.57$ & $+8.61$ & $+15.89$ & $+0.10$ & $+5.55$ & $+16.65$ & $+25.17$ \\
        \addlinespace[0.5pt]
         & \multirow{3}{*}{$\NCins$} & O & 0.00 & 0.10 & 0.38 & 2.30 & 0.00 & 0.10 & 0.29 & 2.39 & 0.00 & 0.10 & 0.57 & 2.11 \\
         &  & E & 0.00 & 3.54 & 22.97 & 36.17 & 0.00 & 0.77 & 9.09 & 18.37 & 0.10 & 5.65 & 17.32 & 27.56 \\
        \rowcolor{gray!15} \cellcolor{white} & \cellcolor{white} & $\Delta$ & $+0.00$ & $+3.44$ & $+22.58$ & $+33.88$ & $+0.00$ & $+0.67$ & $+8.80$ & $+15.98$ & $+0.10$ & $+5.55$ & $+16.75$ & $+25.45$ \\
        \addlinespace[0.5pt]
         & \multirow{3}{*}{$\CS$} & O & 20.34 & 20.57 & 20.97 & 21.73 & 20.34 & 20.40 & 20.59 & 21.01 & 20.30 & 20.87 & 21.81 & 22.38 \\
         &  & E & 18.54 & 14.56 & 14.28 & 16.31 & 17.89 & 14.55 & 14.28 & 15.08 & 17.23 & 12.71 & 13.32 & 15.43 \\
        \rowcolor{gray!15} \cellcolor{white} & \cellcolor{white} & $\Delta$ & $-1.80$ & $-6.01$ & $-6.69$ & $-5.42$ & $-2.45$ & $-5.85$ & $-6.31$ & $-5.93$ & $-3.07$ & $-8.16$ & $-8.49$ & $-6.95$ \\
        \bottomrule
    \end{tabular}
\end{table*}

\begin{figure}[t]
    \centering
    \includegraphics[width=\linewidth]{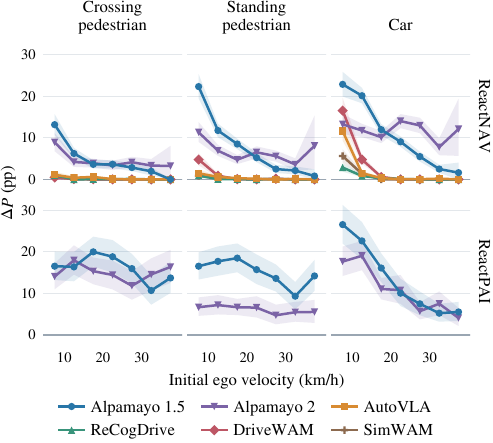}

    \caption{Safety improvement as a function of ego-vehicle speed at the current frame aggregated across all four look-ahead times $\tau$. Shaded areas indicate 95\% bootstrap CIs.}
    \label{fig:deltaP_by_vel}
\end{figure}

\begin{figure*}[t]
    \centering
    \includegraphics[width=\linewidth]{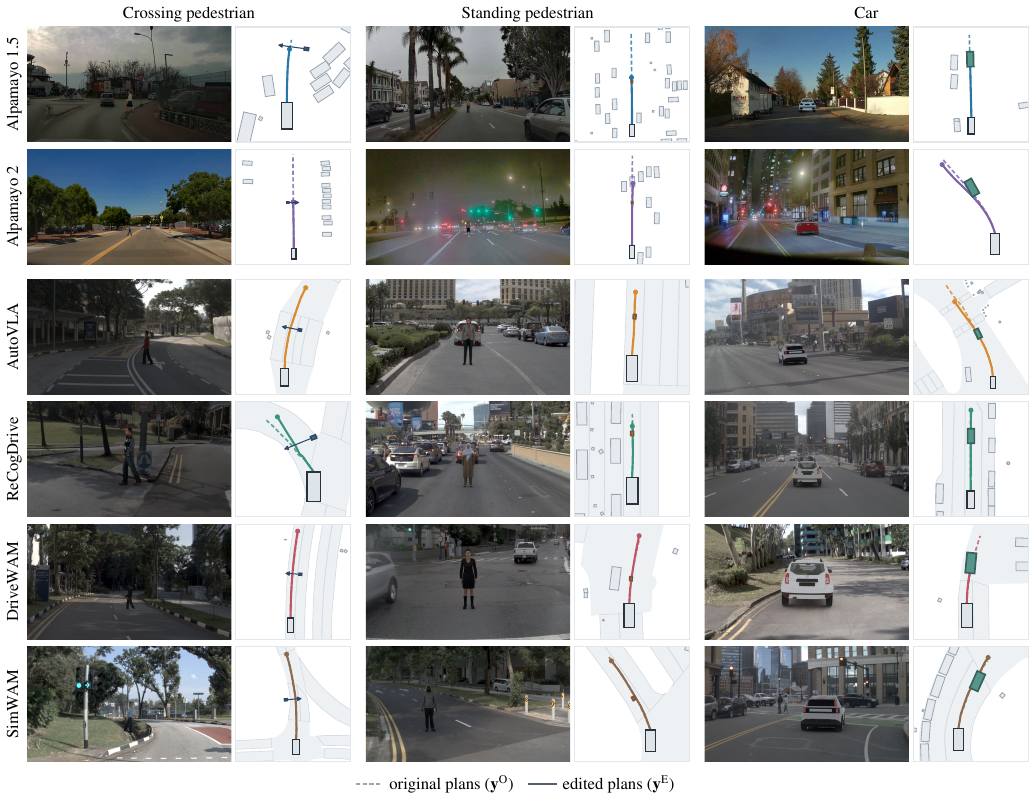}

    \caption{Example predictions at a look-ahead time of $\tau = 2.0\,\mathrm{s}$, one for each model and inserted object: Alpamayo~1.5 and Alpamayo~2 on \gls{reactpai}, and the other four models on \gls{reactnav}. Each example shows the edited front-camera image at the current frame, cropped around the inserted object, next to a bird's-eye view of the edited scene. The arrow marks the crossing pedestrian's motion over the 4\,s horizon. Dashed and solid lines show the model's plans for the original ($\mathbf{y}^{\mathrm{O}}$) and edited ($\mathbf{y}^{\mathrm{E}}$) scenes, respectively, as front-bumper paths over the first 4\,s. Examples are drawn at random from the samples whose original plan collides with the inserted object.}
    \label{fig:trajectory_examples}
\end{figure*}

Additionally, \cref{fig:deltaP_by_vel} shows the safety improvement $\dP$ as a function of the ego-vehicle speed (in km/h) at the current frame, aggregated across all look-ahead times $\tau$. Finally, \cref{fig:trajectory_examples} visualizes the predicted trajectories for some randomly chosen samples from \gls{reactnav} and \gls{reactpai} at a look-ahead time of $\tau = 2.0\,\mathrm{s}$.

\subsection{Unavoidable Collisions}
\label{subsec:supp_unavoidable}
At short look-ahead times, some collisions cannot be avoided even by emergency braking. To quantify their effect, we mark a sample as unavoidable if braking along the ego vehicle's ground-truth path from its measured initial speed, at any constant deceleration of up to $8\,\mathrm{m/s^2}$~\cite{greibe2007braking,8813885}, cannot prevent an at-fault collision with the inserted object within the 4\,s horizon. Collisions are checked as in Appendix~\ref{subsec:supp_metrics_trajectory}, considering only the inserted object, so the criterion does not depend on the evaluated model. Unavoidable samples occur only at $\tau = 0.5\,\mathrm{s}$, where they make up 22\% of the \gls{reactnav} and 28\% of the \gls{reactpai} samples. As \cref{tab:feasible_results} shows, excluding them changes $\dP$ by at most 0.2 percentage points and does not change which improvements are distinguishable from zero.

\begin{table}[t]
    \centering
    \caption{Safety improvement $\dP$ (in percentage points) at $\tau = 0.5\,\mathrm{s}$ on all samples (All, as in \cref{tab:main_results}) and on the avoidable samples only (Avoid.). Unavoidable: share of samples excluded. Gray: the paired 95\% bootstrap CI includes zero.}
    \label{tab:feasible_results}
    \footnotesize
    \setlength{\tabcolsep}{4pt}
    \begin{tabular}{@{}l*{3}{rr}@{}}
        \toprule
        & \multicolumn{2}{c}{Cross.\ ped.} & \multicolumn{2}{c}{Stand.\ ped.} & \multicolumn{2}{c}{Car} \\
        \cmidrule(lr){2-3} \cmidrule(lr){4-5} \cmidrule(l){6-7}
        Model & All & Avoid. & All & Avoid. & All & Avoid. \\
        \midrule
        \multicolumn{7}{l}{\textit{ReactNAV}} \\
        \midrule
        Unavoidable (\%) & \multicolumn{2}{c}{23.0} & \multicolumn{2}{c}{28.3} & \multicolumn{2}{c}{15.8} \\
        \addlinespace[2pt]
        Alpamayo~1.5 & $+0.25$ & $+0.32$ & \textcolor{gray}{$+0.09$} & \textcolor{gray}{$+0.13$} & $+0.12$ & $+0.15$ \\
        Alpamayo~2 & \textcolor{gray}{$+0.06$} & \textcolor{gray}{$+0.08$} & \textcolor{gray}{$+0.03$} & \textcolor{gray}{$+0.04$} & $+0.12$ & $+0.15$ \\
        AutoVLA & \textcolor{gray}{$+0.00$} & \textcolor{gray}{$+0.00$} & \textcolor{gray}{$+0.00$} & \textcolor{gray}{$+0.00$} & \textcolor{gray}{$+0.03$} & \textcolor{gray}{$+0.00$} \\
        ReCogDrive & \textcolor{gray}{$+0.00$} & \textcolor{gray}{$+0.00$} & \textcolor{gray}{$+0.00$} & \textcolor{gray}{$+0.00$} & \textcolor{gray}{$+0.00$} & \textcolor{gray}{$+0.00$} \\
        DriveWAM & \textcolor{gray}{$+0.00$} & \textcolor{gray}{$+0.00$} & \textcolor{gray}{$+0.00$} & \textcolor{gray}{$+0.00$} & \textcolor{gray}{$+0.00$} & \textcolor{gray}{$+0.00$} \\
        SimWAM & \textcolor{gray}{$+0.00$} & \textcolor{gray}{$+0.00$} & \textcolor{gray}{$+0.00$} & \textcolor{gray}{$+0.00$} & \textcolor{gray}{$+0.00$} & \textcolor{gray}{$+0.00$} \\
        \midrule
        \multicolumn{7}{l}{\textit{ReactPAI}} \\
        \midrule
        Unavoidable (\%) & \multicolumn{2}{c}{28.8} & \multicolumn{2}{c}{31.3} & \multicolumn{2}{c}{23.1} \\
        \addlinespace[2pt]
        Alpamayo~1.5 & \textcolor{gray}{$+0.10$} & \textcolor{gray}{$+0.13$} & \textcolor{gray}{$+0.19$} & \textcolor{gray}{$+0.28$} & $+0.67$ & $+0.87$ \\
        Alpamayo~2 & \textcolor{gray}{$+0.00$} & \textcolor{gray}{$+0.00$} & \textcolor{gray}{$+0.00$} & \textcolor{gray}{$+0.00$} & \textcolor{gray}{$+0.10$} & \textcolor{gray}{$+0.12$} \\
        \bottomrule
    \end{tabular}
\end{table}

\subsection{AutoVLA Side Cameras}
\label{subsec:supp_autovla_cameras}
AutoVLA takes three camera views as input, and two sources disagree on which NAVSIM cameras these should be. The AutoVLA paper describes them as the front, front-left, and front-right cameras~\cite{zhou2025autovla}, which in NAVSIM are CAM\_F0, CAM\_L0, and CAM\_R0 (yaw $\approx\pm56^\circ$ for the latter two). The released code, however, uses the side cameras CAM\_L1 and CAM\_R1 (yaw $\approx\pm112^\circ$) as its front-left and front-right views. Throughout the paper, we follow the released code: we keep CAM\_L1/R1 unedited and, for AutoVLA only, exclude the few crossing-pedestrian samples in which the inserted object would enter their field of view (Appendix~\ref{subsec:supp_models}). \Cref{tab:autovla_side_cameras} repeats the evaluation with the configuration described in the paper, \ie, with CAM\_L0/R0 for both the original and the edited scene. The conclusions remain largely the same, with AutoVLA reacting slightly more strongly in the released configuration.

\begin{table*}[t]
    \centering
    \caption{AutoVLA on \gls{reactnav} with CAM\_L1/R1 (released code; used in all other results) or CAM\_L0/R0 as its front-left and front-right views, the latter for both the original and the edited scene. $\dP$ is in percentage points, $\CSC$ in km/h, and $\TIR$ in percent at $\epsilon = 0.05\,\mathrm{m}$, aggregated across all four look-ahead times. Gray: the paired 95\% bootstrap CI includes zero.}
    \label{tab:autovla_side_cameras}
    \footnotesize
    \setlength{\tabcolsep}{4.1pt}
    \begin{tabular}{ll*{4}{r}@{\hspace{16pt}}*{4}{r}@{\hspace{16pt}}*{4}{r}}
        \toprule
        & & \multicolumn{4}{c}{Crossing pedestrian} & \multicolumn{4}{c}{Standing pedestrian} & \multicolumn{4}{c}{Car} \\
        \cmidrule(lr){3-6} \cmidrule(lr){7-10} \cmidrule(lr){11-14}
        Side cameras & Metric & $\tau{=}0.5$ & 1.0 & 1.5 & 2.0 & 0.5 & 1.0 & 1.5 & 2.0 & 0.5 & 1.0 & 1.5 & 2.0 \\
        \midrule
        CAM\_L1/R1 & $\boldsymbol{\dP}$ & \textcolor{gray}{$+0.00$} & \textcolor{gray}{$-0.03$} & $+0.53$ & $+0.90$ & \textcolor{gray}{$+0.00$} & $+0.16$ & $+0.34$ & $+0.96$ & \textcolor{gray}{$+0.03$} & $+0.87$ & $+2.14$ & $+3.26$ \\
         & $\CSC$ & $+0.03$ & \textcolor{gray}{$-0.02$} & \textcolor{gray}{$-0.02$} & \textcolor{gray}{$-0.03$} & \textcolor{gray}{$-0.01$} & $-0.08$ & $-0.16$ & $-0.16$ & $-0.28$ & $-0.59$ & $-0.70$ & $-0.67$ \\
         & $\TIR$ & \multicolumn{4}{c}{34.9} & \multicolumn{4}{c}{19.4} & \multicolumn{4}{c}{17.3} \\
        \addlinespace[2pt]
        CAM\_L0/R0 & $\boldsymbol{\dP}$ & \textcolor{gray}{$+0.00$} & \textcolor{gray}{$+0.03$} & $+0.12$ & $+0.34$ & \textcolor{gray}{$+0.00$} & \textcolor{gray}{$+0.03$} & $+0.16$ & $+0.56$ & \textcolor{gray}{$+0.03$} & $+0.65$ & $+1.83$ & $+2.57$ \\
         & $\CSC$ & $+0.01$ & \textcolor{gray}{$+0.00$} & $-0.04$ & \textcolor{gray}{$-0.01$} & \textcolor{gray}{$-0.01$} & $-0.06$ & $-0.10$ & $-0.10$ & $-0.23$ & $-0.45$ & $-0.59$ & $-0.51$ \\
         & $\TIR$ & \multicolumn{4}{c}{61.4} & \multicolumn{4}{c}{40.8} & \multicolumn{4}{c}{38.2} \\
        \bottomrule
    \end{tabular}
\end{table*}

\subsection{CoT Analysis Details}
\label{subsec:supp_cot_analysis}
This section complements the CoT analysis of \cref{subsec:ablations} with two further experiments: fixed CoTs, which measure the effect of the CoT when the visual input is held fixed, and attention knockout, which tests whether the smaller effect of fixed CoTs on Alpamayo~2 reflects a generally weaker reliance on its CoT.

\subsubsection{Fixed CoTs}
\begin{figure}[t]
    \centering
    \includegraphics[width=\linewidth]{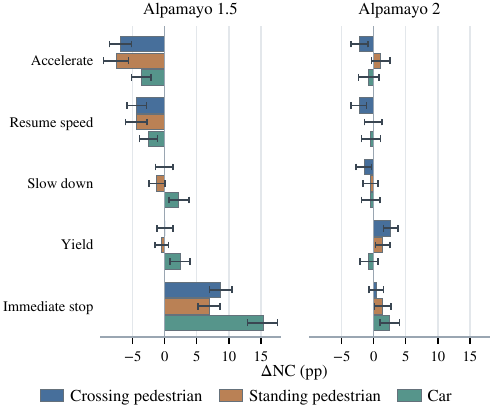}

    \caption{Change in the no-collision rate, $\NC$, of Alpamayo~1.5 and Alpamayo~2 on edited \gls{reactpai} scenes at $\tau = 2.0\,\mathrm{s}$ when their CoT is replaced by a fixed one. Changes are relative to their own generated CoT. Whiskers indicate 95\% bootstrap CIs.}
    \label{fig:fixed_cots}
\end{figure}

On the edited \gls{reactpai} scenes, we replace each Alpamayo model's generated CoT with one of the five predefined CoTs in \cref{tab:fixed_cots}. \Cref{fig:fixed_cots} shows the resulting change in the no-collision rate at $\tau = 2.0\,\mathrm{s}$. For Alpamayo~1.5, instructing the model to accelerate or resume speed produces more unsafe trajectories than letting it generate its own CoT, whereas an immediate-stop instruction increases the no-collision rate by 6.9 to 15.3 percentage points. Even so, it still leaves Alpamayo~1.5 colliding in more than half of the samples. For Alpamayo~2, all changes are at most 2.6 percentage points.

\begin{table}[t]
    \centering
    \caption{Fixed CoTs used in \cref{fig:fixed_cots}. Each fixed CoT replaces the model's generated CoT verbatim, and $\langle$\textit{reason}$\rangle$ is filled in according to the inserted object (bottom).}
    \label{tab:fixed_cots}
    \footnotesize
    \setlength{\tabcolsep}{4pt}
    \begin{tabularx}{\columnwidth}{lX}
        \toprule
        Condition & Fixed CoT \\
        \midrule
        Accelerate & Accelerate since the lane is clear ahead \\
        Resume speed & Resume speed since the lane is clear ahead \\
        Slow down & Slow down due to the $\langle$\textit{reason}$\rangle$ \\
        Yield & Yield due to the $\langle$\textit{reason}$\rangle$ \\
        Immediate stop & Come to a full stop immediately due to the $\langle$\textit{reason}$\rangle$ \\
        \midrule
        Inserted object & $\langle$\textit{reason}$\rangle$ \\
        \midrule
        Crossing pedestrian & pedestrian crossing ahead \\
        Standing pedestrian & pedestrian blocking the lane ahead \\
        Car & vehicle blocking the lane ahead \\
        \bottomrule
    \end{tabularx}
\end{table}

\subsubsection{Attention Knockout}
\label{subsec:supp_attn_knockout}
To test whether the smaller effect of fixed CoTs on Alpamayo~2 (\cref{fig:fixed_cots}) reflects a generally weaker reliance on its CoT, we conduct an attention knockout experiment, in which we hide the CoT tokens from the action expert.

\paragraph{CoT Pathway.}
In both Alpamayo models, each layer of the action expert attends over the action tokens being denoised and over the key--value cache of the corresponding \gls{vlm} layer. This cache contains the prompt, which includes the camera images, followed by the tokens generated by the \gls{vlm}, \ie, the CoT. Since the \gls{vlm} is causal, the cache entries of the prompt do not depend on the CoT. The content of the CoT can therefore affect the predicted trajectory only through the expert's attention to the generated tokens.

\paragraph{Knockout.}
In the knockout condition, we mask the keys of all generated tokens in every layer of the action expert and at every denoising step. Besides the CoT text, these include the special tokens that end the reasoning and start the trajectory, since their key--value entries are computed after the CoT and therefore can also encode information about it. All prompt keys remain visible, including the token that opens the reasoning in Alpamayo~1.5, which is part of the prompt. The attention weights are renormalized over the remaining keys, and the action tokens keep their original rotary position embeddings, so the knockout is equivalent to deleting the generated tokens from the cache without shifting any positions. The model weights and all other computations are unchanged.

\paragraph{Samples and Rollouts.}
We draw 300 samples at random from the 1{,}045 original (unedited) \gls{reactpai} samples and use the same samples for both models. Each model receives the same inputs as in our main \gls{reactpai} evaluation and uses 10 Euler steps of its flow-matching sampler. For each sample, we run three rollouts:
(a) the \emph{natural} rollout $\mathbf{y}$, with the per-sample random seeds of the main evaluation;
(b) the \emph{knockout} rollout $\mathbf{y}^{\mathrm{KO}}$, which uses the same seeds, and hence the same CoT and the same initial noise as $\mathbf{y}$, but hides the CoT from the action expert; and
(c) the \emph{resampled} rollout $\mathbf{y}^{\mathrm{RS}}$, which draws the initial noise of the flow-matching sampler from a different seed but keeps the seed of the text generation, and hence the CoT.
We verified that the three rollouts produce identical CoTs for every sample and that the natural CoT matches the one from the main evaluation for all 300 samples of both models. Since $\mathbf{y}^{\mathrm{KO}}$ shares the initial noise of $\mathbf{y}$, any difference between the two is caused by the CoT alone, and a difference of zero would mean that the CoT has no effect.

\paragraph{Metrics.}
We compare each intervened rollout with the natural one over the first 4\,s of the horizon, expressed in the ego frame at the current frame. We denote by $\mathrm{ADE}^{\mathrm{KO}}$ the \gls{ade} between $\mathbf{y}$ and $\mathbf{y}^{\mathrm{KO}}$ and by $\mathrm{ADE}^{\mathrm{RS}}$ the \gls{ade} between $\mathbf{y}$ and $\mathbf{y}^{\mathrm{RS}}$, and define $\mathrm{FDE}^{\mathrm{KO}}$ and $\mathrm{FDE}^{\mathrm{RS}}$ analogously for the final displacement error at 4\,s. $\mathrm{ADE}^{\mathrm{RS}}$ measures how much the trajectories of each model change under a perturbation that does not involve the CoT, \ie, resampling the initial noise, and thus shows whether one model is generally less sensitive to perturbations than the other; it is not a significance threshold for $\mathrm{ADE}^{\mathrm{KO}}$. All confidence intervals are 95\% percentile bootstrap intervals over samples (10{,}000 resamples), and differences between the two models are computed on the same resamples, \ie, they are paired.

\paragraph{Results.}
\Cref{tab:attn_knockout} reports the full results. Removing the CoT clearly changes the trajectories of both models, so both use their CoT to some extent. However, it moves the trajectory of Alpamayo~2 by 0.33\,m (38\%) less than that of Alpamayo~1.5, and the same pattern holds at the endpoint (FDE of 1.36\,m vs.\ 2.18\,m). In contrast, resampling the initial noise changes the trajectories of both models by the same amount. This confirms that Alpamayo~2 depends less on its CoT, partially explaining the results in \cref{fig:fixed_cots}.

\begin{table}[t]
    \centering
    \caption{Attention knockout on 300 original \gls{reactpai} samples, measured over the first 4\,s. KO: CoT hidden from the action expert, with the same initial noise as the natural rollout. RS: initial noise resampled, CoT unchanged. All values are in meters; means with 95\% bootstrap CIs. Deltas are computed before rounding.}
    \label{tab:attn_knockout}
    \footnotesize
    \setlength{\tabcolsep}{3pt}
    \begin{tabular}{lccc}
        \toprule
        & Alpamayo~1.5 & Alpamayo~2 & Difference \\
        \midrule
        $\mathrm{ADE}^{\mathrm{KO}}$ & 0.89 [0.81, 0.97] & 0.55 [0.50, 0.61] & $+$0.33 [0.27, 0.40] \\
        $\mathrm{ADE}^{\mathrm{RS}}$ & 0.68 [0.61, 0.76] & 0.69 [0.63, 0.75] & $-$0.01 [$-$0.09, 0.09] \\
        $\mathrm{FDE}^{\mathrm{KO}}$ & 2.18 [1.98, 2.38] & 1.36 [1.22, 1.51] & $+$0.82 [0.64, 1.00] \\
        $\mathrm{FDE}^{\mathrm{RS}}$ & 1.96 [1.74, 2.19] & 1.98 [1.80, 2.18] & $-$0.03 [$-$0.28, 0.24] \\
        \bottomrule
    \end{tabular}
\end{table}

\section{Further Training Details}

\subsection{GRPO Exploration Probe}
\label{subsec:supp_rl_probe}
This section details the probe summarized in \cref{sec:teaching_to_react}, which examines whether adding our hazardous edited scenes to ReCogDrive's RL stage would provide a useful learning signal. GRPO can only reinforce behavior that the policy already samples: for each scene, it draws a group of $G$ candidate trajectories and weights each one by its reward normalized within the group, so a group in which all candidates receive the same reward yields no gradient.
 
\paragraph{Setup.}
Starting from ReCogDrive's IL checkpoint, i.e., the initialization of its RL stage, we sample groups of size $G \in \{8, 16, 32\}$ on 300 hazardous scenes built from navtrain (100 per object type, $\tau \in \{1.5, 2.0\}\,\mathrm{s}$), keeping all other RL settings as defined by ReCogDrive~\cite{li2026recogdrive}. $G = 8$ is the group size used in ReCogDrive's RL stage, and the larger groups test whether more exploration would recover a useful signal. We then measure two quantities: the fraction of groups in which the candidates do not all receive the same PDMS reward, i.e., groups that yield a non-zero gradient (Signal), and the fraction of groups that contain at least one safe trajectory, i.e., one with $P = 1$ under the \benchname rules (\cref{sec:reactbench}), whose reward is above the group mean, i.e., groups in which the gradient increases the likelihood of a safe trajectory (Safe signal).
 
\paragraph{Results.}
As shown in \cref{tab:exploration_probe}, even with a group size of 32, only 26.3\% of the groups yield a non-zero gradient, and only 1.3\% contain a trajectory that is safe under \benchname's definition ($P$) and receives an above-average reward, i.e., a safe trajectory that the gradient would actually reinforce. The large gap between the two fractions arises because PDMS often rewards candidates whose simulated motion avoids the pedestrian or only grazes it from the side, even though the planned trajectory itself still drives into it. We therefore conclude that, due to this limited exploration, simply adding our edited scenes to the RL training mix is unlikely to teach the model the desired behavior.
 
\begin{table}[t]
    \centering
    \caption{Percentage of GRPO groups with a non-zero gradient (Signal) and with at least one safe trajectory above the group mean reward (Safe signal) for ReCogDrive's IL checkpoint on 300 hazardous scenes, as a function of group size $G$.}
    \label{tab:exploration_probe}
    \footnotesize
    \setlength{\tabcolsep}{4pt}
    \begin{tabular}{lrrrrrr}
        \toprule
& \multicolumn{3}{c}{Signal (\%)} & \multicolumn{3}{c}{Safe signal (\%)} \\
        \cmidrule(lr){2-4} \cmidrule(lr){5-7}
        Object & $G{=}8$ & $16$ & $32$ & $G{=}8$ & $16$ & $32$ \\
        \midrule
        Crossing pedestrian & 29.0 & 43.0 & 53.0 & 1.0 & 3.0 & 3.0 \\
        Standing pedestrian & 9.0 & 16.0 & 26.0 & 0.0 & 0.0 & 1.0 \\
        Car & 0.0 & 0.0 & 0.0 & 0.0 & 0.0 & 0.0 \\
        \midrule
        All & 12.7 & 19.7 & 26.3 & 0.3 & 1.0 & 1.3 \\
        \bottomrule
    \end{tabular}
\end{table}

\subsection{ReactTrain}
\label{subsec:supp_reacttrain}
This section complements \cref{subsec:sft_on_artificial_targets} with details of the construction of ReactTrain.

\paragraph{Source Samples.}
We draw the source samples only from the training logs of navtrain, following NAVSIM's default split into training and validation logs; navtest, from which \gls{reactnav} and the generalization sets of Appendix~\ref{subsec:supp_generalization_sets} are built, shares no logs with navtrain. As for \gls{reactnav}, we keep only samples with an ego-vehicle speed of at least 5\,km/h at the current frame $t$. We then keep only samples in which the hazardous and the benign variant of each of the three objects pass the placement validity checks of Appendix~\ref{subsec:supp_object_compositing} at all four look-ahead times $\tau$. In addition, both variants must pass a front-camera visibility check: at least 25\% of the area of the object's projected bounding box must lie inside the front-camera image at the current frame $t$, since this is the only camera image that ReCogDrive receives. The benign variants must further satisfy the clearance condition described below. This yields a pool of 8,734 navtrain samples, from which we randomly select 2,000. Of these, 1,957 are rendered successfully in all 24 variants (3 objects $\times$ 2 variants $\times$ 4 look-ahead times) and form the source samples of ReactTrain.

\paragraph{Hazardous Variants.}
The hazardous variants are placed and rendered exactly as in \gls{reactnav}, using the same compositing pipeline, settings and asset pools (Appendix~\ref{subsec:supp_object_compositing}). Since the asset is assigned from the scene identifier, all variants of an object in a source sample, hazardous and benign at every look-ahead time, use the same asset. By construction, the logged ground-truth trajectory collides with the inserted object in every hazardous variant.

\paragraph{Benign Variants.}
The benign variants insert the same object but place it off the ego vehicle's path. As in the hazardous variants, the placement is defined relative to the ego vehicle's front bumper at the look-ahead time $\tau$ on the logged ground-truth trajectory, but the object is shifted by 4\,m perpendicular to the ego vehicle's heading at $\tau$. The standing pedestrian stands at the shifted position and faces the approaching ego vehicle, as in the hazardous variant. The car is the hazardous car shifted by 4\,m and rotated by $180^\circ$ about its center, so that it faces opposite to the ego vehicle's heading. The walking pedestrian does not cross the ego vehicle's path; instead, it walks at 1.4\,m/s along a straight line parallel to the ego vehicle's heading at $\tau$, in the same direction, and passes the shifted position at the current frame $t$. Cars are first placed on the left of the ego vehicle and pedestrians on the right; the other side is used only if the preferred one fails a check at any look-ahead time, so all four look-ahead variants of an object share the same side. A benign placement is valid only if the footprints of the object and the ego vehicle remain at least 1\,m apart along the entire logged future ego trajectory, which we check every 0.05\,s by interpolating between the logged poses. The logged ground-truth trajectory therefore remains safe in every benign variant, which is why we use it as the training target for these samples.

\paragraph{Look-Ahead Times.}
Although all 24 variants are rendered for every source sample, ReactTrain contains each object at only two of the four look-ahead times per source sample. For each object independently, we randomly split the 1,957 source samples into six groups of nearly equal size and assign to each group one of the six pairs of look-ahead times, $\{0.5, 1.0\}$, $\{0.5, 1.5\}$, $\{0.5, 2.0\}$, $\{1.0, 1.5\}$, $\{1.0, 2.0\}$ and $\{1.5, 2.0\}\,\mathrm{s}$. Since each look-ahead time appears in three of the six pairs, each is used for about half of the source samples. The benign variants follow the hazardous ones, \ie, a source sample contributes a benign variant of an object exactly at the two look-ahead times of its hazardous variants. Each source sample thus contributes six hazardous and six benign samples, for a total of 11,742 hazardous and 11,742 benign samples.

\begin{figure*}[t]
    \centering
    \includegraphics[width=\linewidth]{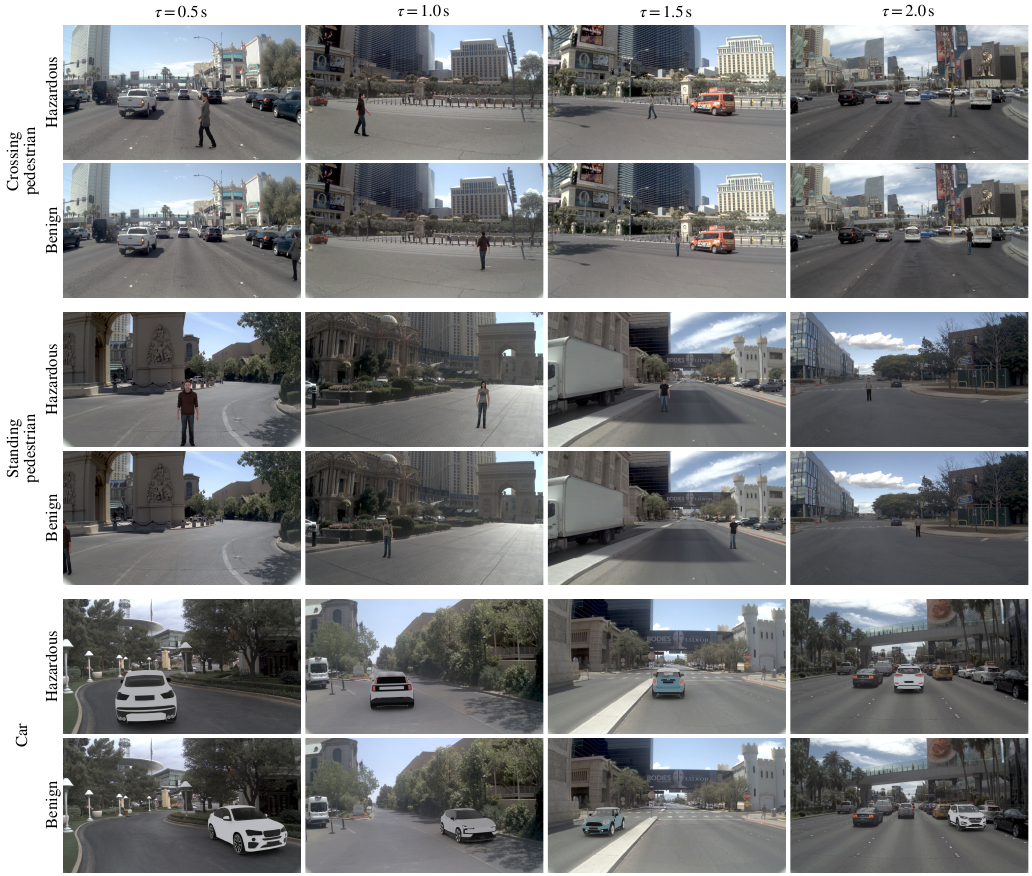}

    \caption{Randomly selected ReactTrain samples for each inserted object and look-ahead time $\tau$, shown in the edited front-camera image at the current frame $t$. Each column shows a different source sample, with its hazardous variant in the top row and its benign variant in the bottom row; both use the same asset.}
    \label{fig:reacttrain_examples}
\end{figure*}

\paragraph{Examples.}
\Cref{fig:reacttrain_examples} depicts the hazardous and benign variants of randomly selected ReactTrain source samples for each inserted object and look-ahead time $\tau$.

\subsection{Fine-Tuning Setup}
\label{subsec:supp_finetuning}
\begin{figure}[t]
    \centering
    \includegraphics[width=\linewidth]{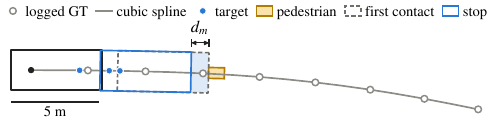}

    \caption{Artificial target trajectory for a ReactTrain sample with a standing pedestrian (orange box) inserted at $\tau = 1.0\,\mathrm{s}$. The ego vehicle (black box) decelerates along a cubic spline fitted to the logged ground-truth waypoints and stops (blue box) a margin $d_{\mathrm{m}}$ before the pose at which it would first contact the pedestrian (dashed box).}
    \label{fig:artificial_target}
\end{figure}

This section complements the experimental details of \cref{subsec:sft_on_artificial_targets}. \Cref{fig:artificial_target} illustrates the construction of the artificial target trajectories that supervise the hazardous ReactTrain samples. We update all parameters of ReCogDrive's action head and keep the \gls{vlm} frozen. Training runs for three epochs with a batch size of 384, which amounts to 276 gradient steps in total. We use AdamW, warming up the learning rate over the first 25 steps to a peak of $3 \times 10^{-5}$ and then decaying it to zero with a cosine schedule. We did not tune these settings, so the reported results reflect what a minimal recipe achieves rather than the best attainable.

\subsection{Generalization Sets}
\label{subsec:supp_generalization_sets}
This section complements \cref{subsec:sft_on_artificial_targets} with details of the two test sets used to evaluate generalization.

\paragraph{Unseen Objects.}
We use five animal assets from the Microsoft Rocketbox library~\cite{10.3389/frvir.2020.561558}: a camel, a horse, a cow, a roe deer, and a sheep. They range from 0.9\,m (roe deer) to 2.1\,m (camel) in height and from 1.25\,m (sheep) to 3.24\,m (camel) in length. As for the other inserted objects, each scene is assigned one animal pseudo-randomly from its identifier with a fixed seed, so all four look-ahead variants of a scene show the same animal. All animals stand in their rest pose with all four hooves on the ground. Each animal is placed at the location of the standing pedestrian, \ie, at the ego vehicle's front-bumper position along its ground-truth trajectory at the look-ahead time $\tau$, but stands side-on, with its body perpendicular to the ego vehicle's heading at $\tau$ and its head pointing to the left or right, chosen pseudo-randomly per scene. Ground estimation, rendering, occlusion handling, and harmonization are identical to those of the other inserted objects (Appendix~\ref{subsec:supp_object_compositing}). For collision checking, we use the ground-plane projection of each asset's bounding box; since the animals are stationary, every contact with them while the ego vehicle is moving is at fault (Appendix~\ref{subsec:supp_metrics_trajectory}).

\begin{figure*}[t]
    \centering
    \includegraphics[width=\linewidth]{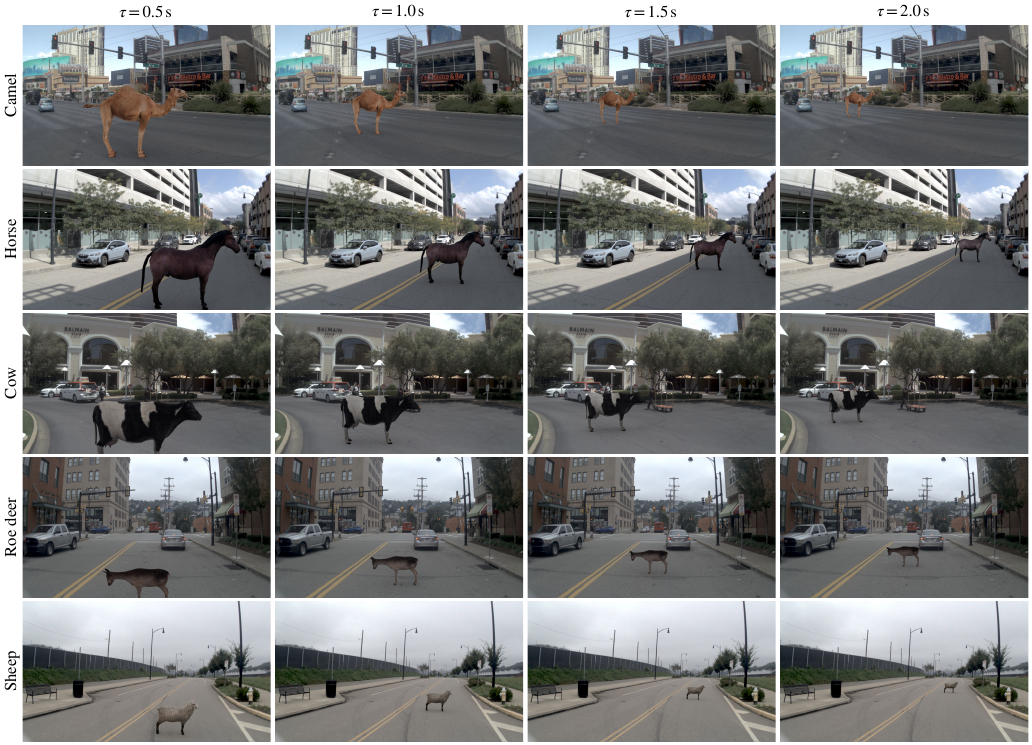}

    \caption{Randomly selected samples of the unseen-objects test set, one per animal, for each look-ahead time $\tau$, shown in the edited front-camera image at the current frame $t$.}
    \label{fig:supp_unseen_objects}
\end{figure*}

We pseudo-randomly select 300 \gls{reactnav} samples and composite an animal into each of them at all four look-ahead times. Compositing fails for one to four samples per look-ahead time, mostly because the rendered animal fails a visibility check or the depth alignment is unreliable. We keep the 291 samples for which all four variants are generated successfully. \Cref{fig:supp_unseen_objects} shows a randomly selected example for each animal.

\paragraph{Real Scenes: Selection.}
For each navtest sample, we determine the ego lane as the lane or lane connector of the nuPlan map under the ego vehicle's rear axle, preferring lanes on the ego vehicle's route and then the one best aligned with its heading. Since map lanes are short segments, we append the lanes that follow it until the resulting lane chain extends at least 60\,m ahead of the ego vehicle. Where a lane splits into several successors, we take the one on the ego vehicle's route or, if the route does not single one out, the one the ego vehicle actually drives into according to its logged trajectory. Distances are measured as arc lengths along the centerline of this lane chain, from the ego vehicle's front bumper to the nearest corner of a road user's bounding box, and the ego vehicle's path is the strip of the ego vehicle's width around this centerline. The target is the nearest vehicle that moves slower than 0.5\,m/s, lies entirely within the lane chain, and is 5--15\,m ahead. We keep a sample if (a) such a target exists, (b) the ego vehicle moves slower than 3\,m/s, (c) no other vehicle, cyclist, or pedestrian, at any speed, lies in the ego vehicle's path closer than the target, and (d) at least half of the target's bounding-box area lies within the path. Of the 12,146 navtest samples, 3,453 satisfy (b), 559 also satisfy (a), 347 also satisfy (c), and 308 also satisfy (d).

\paragraph{Real Scenes: Ego-Status Edit.}
Let $g$ be the distance from the ego vehicle's front bumper to the target along the centerline (5.0--13.7\,m, median 6.4\,m). For each look-ahead time $\tau \in \{0.5, 1.0, 1.5, 2.0\}\,\mathrm{s}$, we set the ego velocity in all four history frames to $v = g / \tau$ along the ego vehicle's heading and the acceleration to zero, and we redraw the past ego trajectory as if the ego vehicle had driven along the centerline at speed $v$, \ie, with consecutive past poses $v \cdot 0.5\,\mathrm{s}$ apart. The camera images and the driving command are left unchanged. The resulting speeds have medians of 46, 23, 15, and 12\,km/h for $\tau = 0.5$, 1.0, 1.5, and 2.0\,s, respectively (ranges 36--99, 18--49, 12--33, and 9--25\,km/h).

\begin{figure*}[t]
    \centering
    \includegraphics[width=\linewidth]{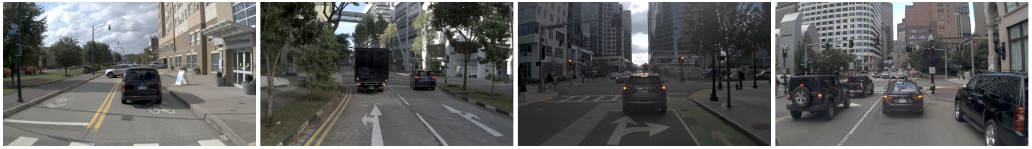}

    \caption{Randomly selected samples of the real-scenes test set, shown in the unedited front-camera image at the current frame $t$. The target is the nearest vehicle ahead in the ego lane.}
    \label{fig:supp_real_scenes}
\end{figure*}

\paragraph{Real Scenes: Scoring.}
Since these scenes have no unedited counterpart, we report absolute values instead of edit effects. $\CS$ is the contact speed (Appendix~\ref{subsec:supp_metrics_trajectory}) at the first at-fault contact with the target, averaged over the predictions with such a contact. Although all targets are nearly stationary at $t$, 54 of them move more than 0.5\,m within the 4\,s horizon, \eg, at the front of a queue that starts moving. We therefore keep the target at its pose at $t$ during scoring, so that a trajectory is not credited for driving towards a vehicle that later moves away; all other road users follow their logged motion. \Cref{tab:generalization_results} reports the mean of each metric over the four look-ahead times. \Cref{fig:supp_real_scenes} shows randomly selected examples.

\end{document}